\documentclass[twoside]{article}

\usepackage[accepted]{aistats2021}

\usepackage[round]{natbib}

\usepackage{url}            
\usepackage{booktabs}       
\usepackage{amsfonts}       
\usepackage{nicefrac}       
\usepackage{microtype}      

\usepackage[svgnames]{xcolor} 
\usepackage[hidelinks=true, hypertexnames=false, colorlinks=true, allcolors=Navy]{hyperref}

\usepackage{amsmath}
\usepackage{amssymb}                                                            
\usepackage{algorithm}                                                          
\usepackage{algorithmic} 
\usepackage{graphicx}

\newcommand \indep{\mathop{\perp\!\!\!\!\perp}}

\usepackage{bm}
\usepackage{tabularx}
\usepackage{multirow}
\usepackage{booktabs}
\usepackage{pifont}
\usepackage{bbm}

\usepackage{mathtools}
\usepackage{wrapfig}
\usepackage{refcount}
\usepackage{multirow}

\def\rnum#1{\resizebox{0.5em}{\height}{\expandafter{\romannumeral #1}}}
\def\Rnum#1{\resizebox{0.5em}{\height}{\uppercase\expandafter{\romannumeral #1}}}
\newcommand \myra{i}
\newcommand \myrb{i\hspace{-1pt}i}
\newcommand \myrc{i\hspace{-1pt}i\hspace{-1pt}i}
\newcommand \myrd{i\hspace{-1pt}v}

\newtheorem{definition}{Definition}
\newtheorem{theorem}{Theorem}

\newtheorem{assumption}{Assumption}

\renewcommand{\algorithmicrequire}{\textbf{Input:}}
\renewcommand{\algorithmicensure}{\textbf{Output:}}

\makeatletter
\let\MYcaption\@makecaption
\makeatother

\usepackage{subcaption}

\makeatletter
\let\@makecaption\MYcaption
\makeatother

\DeclareMathOperator{\E}{\mathbb{E}}

\DeclareMathOperator{\pr}{\mathrm{P}}
\DeclareMathOperator{\I}{\mathbf{1}}

\newcommand{\pzz}{p_{00}}
\newcommand{\pzo}{p_{01}}
\newcommand{\poz}{p_{10}}
\newcommand{\poo}{p_{11}}
\newcommand{\mz}{\alpha}
\newcommand{\mo}{\beta}

\newcommand{\Yz}{Y_{A \Leftarrow 0}}
\newcommand{\Yo}{Y_{A \Leftarrow 1 \parallel \pi}}

\newcommand{\ePz}{\hat{p}^{A \Leftarrow 0}_{\theta}}
\newcommand{\ePo}{\hat{p}^{A \Leftarrow 1 \parallel \pi}_{\theta}}
\newcommand{\eLz}{\hat{l}^{A \Leftarrow 0}_{\theta}}
\newcommand{\eLo}{\hat{l}^{A \Leftarrow 1 \parallel \pi}_{\theta}}
\newcommand{\eUz}{\hat{u}^{A \Leftarrow 0}_{\theta}}
\newcommand{\eUo}{\hat{u}^{A \Leftarrow 1 \parallel \pi}_{\theta}}

\usepackage{cleveref}
\crefname{assumption}{assumption}{assumptions}

\allowdisplaybreaks

\newcommand{\arefex}{\Cref{subsec-a:rpo}}
\newcommand{\arefsem}{\Cref{subsubsec-a:sem}}
\newcommand{\arefpse}{\Cref{subsec-a:pse}}
\newcommand{\arefasmp}{\Cref{sec-a:asmp}}
\newcommand{\arefth}{\Cref{sec-a:th1}}
\newcommand{\arefhuber}{\Cref{sec-a:mp_huber}}
\newcommand{\arefconv}{\Cref{sec-a:convergence}}

\newcommand{\arefcomp}{\Cref{sec-a:comparison}}
\newcommand{\arefextend}{\Cref{sec-a:extension}}
\newcommand{\arefexpset}{\Cref{subsec-a:set}}
\newcommand{\arefsynthd}{\Cref{subsubsec-a:synthdata}}

\newcommand{\arefsynthp}{\Cref{subsubsec-a:synthpscf}}
\newcommand{\arefreald}{\Cref{subsubsec-a:realdata}}

\newcommand{\arefaddexp}{\Cref{sec-a:addexp}}
\newcommand{\areflogisticexp}{\Cref{subsec-a:logisticexp}}
\newcommand{\areferrbexp}{\Cref{subsec-a:errbexp}}
\newcommand{\arefadmexp}{\Cref{subsec-a:admexp}}
\newcommand{\arefboundexp}{\Cref{subsec-a:boundexp}}
\newcommand{\arefextendedexp}{\Cref{subsec-a:extendedexp}}

\usepackage{sectsty}
\usepackage{titlecaps}
\Addlcwords{is for an of and on with in via} 
\sectionfont{\uppercase}
\subsectionfont{\titlecap}
\subsubsectionfont{\titlecap}

\begin{document}{

\twocolumn[

\aistatstitle{Learning Individually Fair Classifier \\
	with Path-Specific Causal-Effect Constraint}
\aistatsauthor{ Yoichi Chikahara$^{1,3}$ \And Shinsaku Sakaue$^2$ \And  Akinori Fujino$^1$ \And Hisashi Kashima$^3$ }

\aistatsaddress{$^1$NTT  \And $^2$The University of Tokyo \And $^3$Kyoto University}
]

\begin{abstract}
	Machine learning is used to make decisions for individuals in various fields, which require us to achieve good prediction accuracy while ensuring fairness with respect to sensitive features (e.g., race and gender). This problem, however, remains difficult in complex real-world scenarios. To quantify unfairness under such situations, existing methods utilize {\it path-specific causal effects}.  However, none of them can ensure fairness for each individual without making impractical functional assumptions about the data. In this paper, we propose a far more practical framework for learning an individually fair classifier. To avoid restrictive functional assumptions, we define the {\it probability of individual unfairness} (PIU) and solve an optimization problem where PIU's upper bound, which can be estimated from data, is controlled to be close to zero. We elucidate why our method can guarantee fairness for each individual. Experimental results show that our method can learn an individually fair classifier at a slight cost of accuracy.
\end{abstract}
	
	\section{Introduction} \label{sec1}
	
	Machine learning is increasingly being used to make critical decisions that severely affect people's lives (e.g., loan approvals \citep{khandani2010consumer}, hiring decisions \citep{houser2019can}, and recidivism predictions \citep{machinebias}). The huge societal impact of such decisions on people's lives raises concerns about fairness because these decisions may be discriminatory with respect to {\it sensitive features}, including race, gender, religion, and sexual orientation. 
	
	Although many researchers have studied how to make fair decisions while achieving high prediction accuracy \citep{dwork2012fairness,feldman2015certifying,hardt2016equality}, it remains a challenge in complex real-world scenarios. For instance, consider hiring decisions for physically demanding jobs. Although it is discriminatory to reject applicants based on gender, since the job requires physical strength, it is sometimes {\bf not} discriminatory to reject the applicants due to physical strength. Since physical strength is often affected by gender, rejecting applicants due to physical strength leads to gender difference in the rejection rates. Although this difference due to physical strength is {\bf not} always unfair, it is removed when using traditional methods (e.g., \cite{feldman2015certifying}). Consequently, even if there is a man who has a much more physical strength than a woman, these methods might reject him to accept her, which severely reduces the prediction accuracy.
	
	To achieve high prediction accuracy, we need to remove only an unfair difference in decision outcomes. To measure this difference, existing methods utilize a path-specific causal effect \citep{avin2005identifiability}, which we call an {\it unfair effect}. Using unfair effects, the {\it path-specific counterfactual fairness} (PSCF) method \citep{chiappa2018path} aims to guarantee fairness for each individual; however, achieving such an individual-level fairness is possible only when the data are generated by a restricted class of functions. By contrast, {\it fair inference on outcome} (FIO) \citep{nabi2018fair} does not require such demanding functional assumptions; however, it cannot ensure individual-level fairness.
	
	The goal of this paper is to propose a learning framework that guarantees individual-level fairness without making impractical functional assumptions. For this goal, we train a classifier by forcing the {\it probability of individual unfairness} (PIU), defined as the probability that an unfair effect is non-zero, to be close to zero. This, however, is difficult to achieve because we cannot estimate PIU from data. To overcome this difficulty, we derive its upper bound that can be estimated from data and solve a penalized optimization problem where the upper-bound value is controlled to be close to zero.

	{\bf Our contributions} are summarized as follows:
	\setlength{\leftmargini}{5pt}
	\begin{itemize}
		\item We establish a framework that guarantees fairness for each individual without restrictive functional assumptions on the data (\Cref{table0}). To achieve this, we make the PIU value close to zero by imposing a penalty that reduces its upper bound value, which can be estimated from data.
		\item  We elucidate why imposing such a penalty guarantees individual-level fairness in \Cref{subsubsec:prop-obj,subsec:feasible}. We also show how our method can be extended to address cases where there are unobserved variables called {\it latent confounders} in \Cref{subsec:extension}.
		\item We experimentally show that our method makes much fairer predictions for each individual than the existing methods at a slight cost of prediction accuracy.
	\end{itemize}
	
		\begin{table}[t] 
		\centering
		\caption{Comparison with existing methods}
		\label{table0}
		\scalebox{0.88}{
			\tabcolsep=1.5mm
			\begin{tabular}{lcc}
				\toprule
				Method & Individually fair & Functional assumptions \\		
				\midrule						
				Our method &     Yes &   Unnecessary  \\
				PSCF  & Yes & Necessary\\
				FIO   &     	No  &  Unnecessary \\
				\bottomrule
			\end{tabular}
		}
	\end{table}

	\section{Preliminaries} \label{sec2} \label{sec:background}

	\subsection{Problem statement} \label{subsec:problem}
	
	In this paper, we consider a binary classification task. We train classifier $h_{\theta}$ with parameter $\theta$ to predict decision outcome $Y \in \{0, 1\}$ from the features of each individual $\textbf{\textit{X}}$, which contains sensitive feature $A \in \{0, 1\}$. 
	
	We seek classifier parameter $\theta$ that achieves a good balance between prediction accuracy and fairness with respect to sensitive feature $A$. Suppose that we have loss function $L_{\theta}$ and penalty function $G_{\theta}$, which respectively measure prediction errors and unfairness based on $\theta$. Formally, given $n$ training instances $\{(\textbf{\textit{x}}_i, y_i)_{i=1}^n\}$, our learning problem is formulated as follows:
	\begin{align}
	\underset{\theta}{\text{min}}
	\quad &\frac{1}{n}\sum_{i=1}^n L_{\theta}(\textbf{\textit{x}}_i, y_i) + \lambda G_{\theta} (\textbf{\textit{x}}_1, \dots, \textbf{\textit{x}}_n),
	\label{eq-Opt3}
	\end{align}		
	where $\lambda \ge 0$ is a hyperparameter.

	To achieve a high prediction accuracy, penalty function $G_{\theta}$ must be designed such that we can avoid imposing unnecessary penalizations. To do so, we utilize a {\it causal graph}, which is a directed acyclic graph (DAG) whose nodes and edges represent random variables and causal relationships, respectively \citep{pearl2009causality}. We assume that a causal graph is provided by domain experts or can be inferred from data \citep{glymour2019review}; this assumption is common in many existing methods \citep{chiappa2018path,kusner2017counterfactual,nabi2018fair,zhang2016causal}. 
	
	As an example of a causal graph, consider a scenario for hiring decisions for a physically demanding job. In this scenario, a causal graph might be given, as shown in \Cref{fig1}(a), where $A, Q, D, M \in \textbf{\textit{X}}$ represent gender, qualifications, the number of children, and physical strength, respectively. This graph expresses our knowledge that prediction $Y$ is unfair only if it is based on gender $A$. To do so, we regard direct pathway $A$ $\rightarrow$ $Y$ as unfair pathway $\pi$ (i.e., $\pi = \{A \rightarrow Y\}$). 
	
	In fact, in the above simple case, we can naively remove the unfairness by making a prediction without gender $A$; however, this is insufficient when we consider a complex scenario with multiple unfair pathways. 
	
	For instance, as shown in \Cref{fig1}(b), we may regard not only $A \rightarrow Y$ but also the pathway through the number of children $D$ ($A \rightarrow D \rightarrow Y$) as unfair (because it is also discriminatory to reject women because of the possibility of bearing children). In this case, a naive approach to ensure fairness is to predict without $A$ or $D$. This, however, might unnecessarily decrease the prediction accuracy. For instance, consider a case where number of children $D$ is only slightly affected by gender $A$ (e.g., the applicants had gender-equitable opportunities to take parental leave in the past) and largely influenced by other unobserved features that are important for prediction (e.g., communication skills). Then predicting without $D$ will seriously decrease the accuracy while contributing almost nothing to fairness.

	To address such cases, given unfair pathways $\pi$, we design penalty function $G_{\theta}$ by quantifying the unfairness based on data. To do so, we utilize path-specific causal effects, which are described in the next section.

	\begin{figure}[t]
		\centering
		\includegraphics[height=2.25cm]{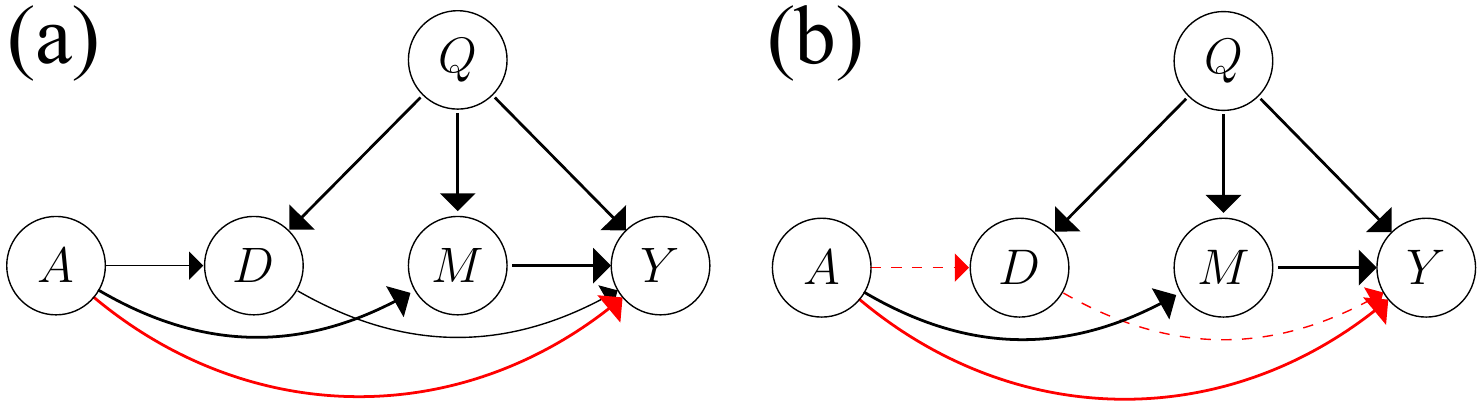}
		\caption{Causal graphs representing a scenario of hiring decisions for physically demanding jobs: Unfair pathways are (a): red solid edge $A \rightarrow Y$; (b): $A \rightarrow Y$ and red dashed pathway $A \rightarrow D \rightarrow Y$.}
		\label{fig1}
	\end{figure}

	\subsection{Path-specific causal effects}  \label{subsec:pse}
	
	A path-specific causal effect measures how largely an observed variable influences another variable via pathways in a causal graph \citep{avin2005identifiability}. Although prediction $Y$ is not observed but is given by classifier $h_{\theta}$, we can utilize this measure to quantify the influence of sensitive feature $A$ on $Y$ via unfair pathways $\pi$. 
	
 	With a path-specific causal effect, this influence is measured by the difference of the two predictions, which is obtained by modifying input features $\textbf{\textit{X}}$. To illustrate these predictions, consider a case where sensitive feature $A$ is gender. Then for each woman ($A=0$), one prediction is made by directly taking her attributes as input, and another is made with {\it counterfactual} attributes, which would be observed if she were male ($A=1$); for each man, these predictions are made using the counterfactual attributes that would be if he were female (see, \arefex\ for details). Although such counterfactual attributes are not observed, they can be computed by a {\it structural equation model} (SEM).

	An SEM consists of {\it structural equations}, each of which expresses variable $V \in \{\textbf{\textit{X}}, Y\}$ by deterministic function $f_V$ \citep{pearl2009causality}. Each function $f_V$ takes as input two types of variables. One is observed variables, which are the parents of $V$ in a causal graph, and the other is unobserved noise $U_V$, which expresses random variable $V$ using deterministic function $f_V$. 
	
	For instance, structural equations over $D, M \in \textbf{\textit{X}}$ in the causal graph in \Cref{fig1}(b) may be formulated as 
		\begin{align}
			\label{causalmodel1}
		\begin{aligned}
			&D= f_D(A, Q, U_D) = A + U_D Q, \\
			&M = f_M(A, Q, U_M) = 3 A + 0.5 Q + U_M,  
		\end{aligned} 
		\end{align}
	where $U_D$ is multiplicative noise and $U_M$ is additive noise. By contrast, the structural equation over prediction $Y$ is formulated using classifier $h_{\theta}$. If $h_{\theta}$ is deterministic, it is expressed as $Y = h_{\theta}(A, Q, D, M)$; otherwise, $Y = h_{\theta}(A, Q, D, M, U_Y)$, where $U_Y$ is a random variable used in the classifier. See \arefsem\ for a formal definition of SEM in our setting.

		Structural equations \eqref{causalmodel1} can be used to compute the (counterfactual) attributes of $D$ and $M$ that are observed when $A = a$ ($a \in \{0, 1\}$) as
			\begin{align}
					D(a) = a + U_D Q, \quad M(a) = 3a + 0.5 Q + U_M. \label{pm}
				\end{align}
			If \eqref{pm} is available, we can obtain attributes $D(0)$, $D(1)$, $M(0)$, and $M(1)$ for each individual. 
			
			Using these attributes, we can compute a path-specific causal effect for each individual, which we call an unfair effect. For instance, when measuring the influence via unfair pathways $\pi = \{A \rightarrow Y, A \rightarrow D \rightarrow Y\}$ in \Cref{fig1}(b), we define an unfair effect as the difference of two predictions $\Yo - \Yz$, where $\Yz$ and $\Yo$ are called {\it potential outcomes} and given as
					\begin{align}
						\label{po}
						\begin{aligned}
							&\Yz = h_{\theta}(0, Q, D(0), M(0)), \\
							&\Yo = h_{\theta}(1, Q, D(1), M(0)).
						\end{aligned}
					\end{align}
		In \eqref{po}, the inputs of $\Yz$ are $A=0$, $D(0)$, and $M(0)$, all of which are given using the same value, $a=0$. By contrast, the inputs of $\Yo$ are formulated based on unfair pathways $\pi$; we use the value $a=1$ {\bf only} for $A$ and $D$ (i.e., $A=1$ and $D(1)$), which correspond to the nodes on $\pi = \{A \rightarrow Y, A \rightarrow D \rightarrow Y\}$ (see \arefpse\  for the formal definition).\footnote{We can also consider different potential outcomes $Y_{A\Leftarrow1}$ and $Y_{A\Leftarrow0 \parallel \pi}$, where all inputs of $Y_{A\Leftarrow1}$ are given using the value $a=1$, and $Y_{A\Leftarrow0 \parallel \pi}$ is formulated using $a=0$ only for the inputs that correspond to the nodes on pathways $\pi$.}
			

	In practice, however, we cannot compute an unfair effect for each individual. This is because we cannot formulate an SEM since it requires a deep understanding of true data-generating processes; consequently, for instance, we can obtain $D(a)$ and $M(a)$  in \eqref{pm} {\bf only} for either $a=0$ or $a=1$ but {\bf not both}. Due to this issue, existing methods use the (conditional) expected values of unfair effects, which can be estimated from data.

	\section{Existing methods and their weaknesses} \label{subsec:weak-existing}

Using unfair effects, two types of existing methods have been proposed. Unfortunately, as presented in \Cref{table0}, each has a weakness. One requires restrictive functional assumptions, and the other cannot ensure individual-level fairness. Below we describe their details.

	\subsection{Methods for ensuring individual-level fairness}  \label{subsec:weak-existing1}
	
	The PSCF method \citep{chiappa2018path} aims to satisfy the following individual-level fairness criterion:
\begin{definition}[\citet{wu2019pc}]
	\label{def_indiv}
	Given unfair pathways $\pi$ in a causal graph, classifier $h_{\theta}$ achieves a \textbf{(path-specific) individual-level fairness} if 
	\begin{align}
		\E_{\Yz, \Yo}[ \Yo  - \Yz | \textbf{\textit{X}} = \textbf{\textit{x}} ] = 0 \label{eq-CE}
	\end{align}
	holds for any value of $\textbf{\textit{x}}$ of input features $\textbf{\textit{X}}$.
\end{definition}
Condition \eqref{eq-CE} states that classifier $h_{\theta}$ is individually fair if the {\it conditional mean unfair effect} is zero, which is an average over individuals who have identical attributes for all features in $\textbf{\textit{X}}$. Since $\Yz$ and $\Yo$ are expressed using classifier parameter $\theta$ as shown in \eqref{po}, we need to find appropriate $\theta$ values to satisfy \eqref{eq-CE}.

	Unfortunately, such $\theta$ values can be found only in restricted cases. As pointed out by \citet{wu2019pc}, this is because we cannot always estimate the conditional mean unfair effect in \eqref{eq-CE}. For instance, when potential outcomes are given as $\eqref{po}$, since the conditional mean unfair effect is conditioned on $A$ and $D$, estimating it requires the joint distribution of $D(0)$ and $D(1)$. This joint distribution, however, is unavailable because we cannot jointly obtain them as explained in \Cref{subsec:pse}.
	
	Due to this issue, the PSCF method (and the one in \citep[Section S4]{kusner2017counterfactual}) can achieve individual-level fairness only when the data are generated from a restricted functional class of SEMs. Specifically, these existing methods assume that each variable $V$ follows an additive noise model $V = f_V(\textbf{\textit{pa}}(V)) + U_V$, where $\textbf{\textit{pa}}(V)$ denotes the parents of $V$ in the causal graph. Unfortunately, this model cannot express data-generating processes in many cases. For instance, it cannot express variable $D$ in \eqref{causalmodel1} due to multiplicative noise $U_D$. Traditionally, this assumption has been used to infer causal graphs \citep{hoyer2009nonlinear,shimizu2006linear}. However, as mentioned in \citet{glymour2019review}, since more recent causal graph discovery methods require much weaker assumptions \citep{zhang2009identifiability,stegle2010probabilistic}, the presence of such an assumption severely restricts the scope of their applications.
	
	\subsection{Another method for removing unfair effects}  \label{subsec:weak-existing2}

		To avoid the aforementioned restrictive functional assumption, the FIO method \citep{nabi2018fair} aims to remove the {\it mean unfair effect} over \textbf{all} individuals, which is expressed as
	\begin{align}
		\label{eq-PE}
		\begin{aligned}
		&\E_{\Yz, \Yo}[ \Yo  - \Yz  ] \\
		&= \pr(\Yo = 1) - \pr(\Yz = 1).
		\end{aligned}
	\end{align}
		In \eqref{eq-PE}, marginal probabilities $\pr(\Yz=1)$ and $\pr(\Yo=1)$ can be estimated under much weaker assumptions than the conditional mean unfair effect in \eqref{eq-CE}. We detail these assumptions in \arefasmp\ and how to derive the estimators in \arefhuber.

	However, removing this mean unfair effect does not imply individual-level fairness. This is because depending on input features $\textbf{\textit{X}}$, unfair effects might be largely positive for some individuals and largely negative for others, which is seriously discriminatory for these individuals. Note that we cannot resolve this issue simply using e.g., the mean of the absolute values of the unfair effects. This is because estimating such a quantity requires a joint distribution of $\Yz$ and $\Yo$; however, this joint distribution is unavailable because we cannot obtain both $\Yz$ and $\Yo$ for each individual without an SEM.

	\section{Proposed method}
	
	\subsection{Overcoming weaknesses of existing methods}
	
	To resolve the weaknesses of the existing methods, we propose a framework that guarantees individual-level fairness without restrictive functional assumptions. 
	
	For this goal, we aim to train a classifier by forcing an unfair effect to be zero for {\bf all} individuals: i.e., making potential outcomes take the same value (i.e., $\Yz = \Yo = 0$ or $\Yz = \Yo = 1$) with probability $1$ regardless of the values of input features $\textbf{\textit{X}}$. This is sufficient to satisfy the individual-level fairness condition (\Cref{def_indiv}) because it restricts the potential outcome values more severely than the latter condition, where potential outcomes $\Yz = \Yo$ can take $0$ or $1$ depending on $\textbf{\textit{X}}$'s values. Although such a fairness condition may be overly severe and might decrease the prediction accuracy, in \Cref{sec:experiment} we experimentally show that our method can achieve comparable accuracy to the existing method for ensuring individual-level fairness (i.e., the PSCF method \citep{chiappa2018path}). 
	
	Compared with PSCF, our method has a clear advantage in that it requires much weaker assumptions. We only need to estimate the marginal potential outcome probabilities in \eqref{eq-PE}, which only requires several conditional independence relations and the graphical condition on unfair pathways $\pi$ (see \arefasmp\ for our assumptions). Furthermore, we can relax these assumptions to address some cases where there are unobserved variables called latent confounders (\Cref{subsec:extension}).
	
	\subsection{Achieving individual-level fairness with PIU}\label{subsec:piu}
	
	We aim to make potential outcomes take the same value for all individuals. To this end, we formulate penalty function $G_{\theta}$ based on the following quantity:
	\begin{definition}
		For unfair pathways $\pi$ in a causal graph and potential outcomes $\Yz, \Yo \in \{0, 1\}$, we define the \textbf{probability of individual unfairness (PIU)} by $\pr(\Yz \neq \Yo)$. 
	\end{definition}
	Intuitively, PIU is the probability that potential outcomes $\Yz$ and $\Yo$ take different values. 
	
	Unlike the conditional mean unfair effect in \Cref{def_indiv}, PIU is not conditioned on features $\textbf{\textit{X}}$ of each individual.
	
	Nonetheless, PIU can be used to guarantee individual-level fairness. By constraining PIU to zero, we can guarantee that potential outcomes take the same value (i.e., $\Yz = \Yo = 0$ or $\Yz = \Yo = 1$) with probability $1$ regardless of the values of $\textbf{\textit{X}}$, which is sufficient to ensure individual-level fairness. 
	
	Unfortunately, we cannot directly impose constraints on PIU. This is because estimating the PIU value requires the joint distribution of $\Yz$ and $\Yo$, which is unavailable as described in \Cref{subsec:weak-existing2}.
	
	To overcome this issue, instead of PIU, we utilize its upper bound that can be estimated from data. Specifically, to make the PIU value close to zero, we formulate a penalty function that forces the upper bound on PIU to be nearly zero, which is described in the next section.

	\subsection{Penalty by upper bound on PIU}
	
	\subsubsection{Upper bound formulation} \label{subsubsec:ub}
	
	To make the PIU value small, we utilize the following upper bound on PIU:
	\begin{theorem}[Upper bound on PIU]\label{th1}
		Suppose that potential outcomes $\Yz$ and $\Yo$ are binary. Then for any joint distribution of potential outcomes $\pr(\Yz, \Yo)$, PIU is upper bounded as follows:
		\begin{align}
		\pr(\Yz \neq \Yo) \leq 2 \pr^I(\Yz \neq \Yo),
		\label{eq-Th1} 
		\end{align}
		where $\pr^I$ is an independent joint distribution, i.e., $\pr^I(\Yz, \Yo)$ $=$ $\pr(\Yz) \pr(\Yo)$.
	
	\end{theorem}
	The proof is detailed in \arefth. \Cref{th1} states that whatever joint distribution potential outcomes $\Yz$ and $\Yo$ follow, the resulting PIU value is at most twice the PIU value that is approximated with independent joint distribution $\pr^I$. 
	
	Note that this upper bound can be larger than $1$, and if so, the PIU value is not controlled because PIU is at most $1$. However, since PIU is always smaller than its upper bound, by making the upper bound close to zero, we can guarantee that PIU is also close to zero. 
	
		\subsubsection{Estimating upper bound} \label{subsubsec:ub_es}
	
	 	Using the observed data, we estimate the upper bound on PIU in \eqref{eq-Th1}, which is twice the value of the approximated PIU. Recall that this approximated PIU is the probability that potential outcomes $\Yz$ and $\Yo$ take different values when they are independent. Since potential outcomes are binary, it is expressed as the probability that potential outcome values are $(\Yz, \Yo)$ $=$ $(0,1)$ or $(1,0)$; in other words,
	\begin{align}
		\begin{aligned}
		&\pr^I(\Yz \neq \Yo)  \\
		= &\pr(\Yo = 1) (1 - \pr(\Yz = 1) )  \\
		&+ (1 - \pr(\Yo=1)  ) \pr(\Yz = 1). 
		\end{aligned}\label{eq-Th1-s}
		\end{align}

	Various estimators can be used to estimate marginal probabilities $\pr(\Yz=1)$ and  $\pr(\Yo=1)$ in \eqref{eq-Th1-s}. Among them, we utilize the computationally efficient estimator in \cite{huber2014identifying}, which can be computed in $O(n)$ time, where $n$ is the number of training instances. 
	
	Let $c_{\theta}(\textbf{\textit{X}}) = \pr(Y=1 | \textbf{\textit{X}} )$ denote the conditional distribution provided by classifier $h_{\theta}$; we let $c_{\theta}(\textbf{\textit{X}}) = h_{\theta}(\textbf{\textit{X}}) \in \{0, 1\}$ if $h_{\theta}$ is a deterministic classifier. For instance, suppose that the causal graph is given as shown in \Cref{fig1}(b) and that the features of $n$ individuals are provided as $\{\textbf{\textit{x}}_i\}_{i=1}^n$ $=$ $\{a_i, q_i, d_i, m_i\}_{i=1}^n$. Then $\pr(\Yz=1)$ and  $\pr(\Yo=1)$ can be estimated as the following weighted averages:
	\begin{align}
		\begin{aligned}
	&\ePz = \frac{1}{n} \sum_{i=1}^n \I(a_i = 0) \hat{w}_i c_{\theta}(a_i, q_i, d_i, m_i) \  \mbox{and}\\
	&\ePo = \frac{1}{n} \sum_{i=1}^n \I(a_i = 1) \hat{w}_i' c_{\theta}(a_i, q_i, d_i, m_i),
		\end{aligned}
		\label{mp_huber}
	\end{align}
	where $\I(\cdot)$ is an indicator function, and $\hat{w}_i$ and $\hat{w}_i'$ are the following weights for individual $i \in \{1, \dots, n\}$: 
	\begin{align*}
		&\hat{w}_i = \frac{1}{\hat{\pr}(A=0 | q_i)}, \\
		&\hat{w}_i' = \frac{\hat{\pr}(A=1 | q_i, d_i) \hat{\pr}(A=0 | q_i, d_i, m_i)}{\hat{\pr}(A=1 | q_i) \hat{\pr}(A=0 | q_i, d_i) \hat{\pr}(A=1 | q_i, d_i, m_i)},
	\end{align*}
 where $\hat{\pr}$ is a conditional distribution, which we infer in the same way as \citet{zhang2018equality}, i.e., by learning a statistical model (e.g., a neural network) from the training data beforehand.\footnote{Note that FIO infers conditional distributions not by learning statistical models beforehand but by simultaneously learning them with the predictive model of $Y$ \citep{nabi2018fair}. This is because unlike our method, it addresses not only training a classifier but also learning a generative model of joint distribution $\pr(\textbf{X}, Y)$.} We derive the estimators \eqref{mp_huber} in \arefhuber. 
 
 In \eqref{mp_huber}, marginal probabilities $\ePz$ and $\ePo$ are estimated by taking a weighted average of conditional probability $c_{\theta}$ over individuals with $A=0$ and $A=1$, respectively, which is a widely used estimation technique called {\it inverse probability weighting} (IPW).

	\subsubsection{Formulating penalty function} \label{subsubsec:prop-obj}
	
	To learn an individually fair classifier, we force the estimated value of the upper bound on PIU to be close to zero by minimizing the objective function \eqref{eq-Opt3} with the following penalty function $G_{\theta}$:
	\begin{align}
		&G_{\theta}(\textbf{\textit{x}}_1, \dots \textbf{\textit{x}}_n) \nonumber \\
		&=  \ePo (1 - \ePz) + (1 - \ePo) \ePz.
		\label{eq-penalty}
	\end{align}
	For instance, if $\ePz$ and $\ePo$ are given as the weighted estimators \eqref{mp_huber},
	to reduce the value of $G_{\theta}$, our method imposes strong penalties on the predictions for individuals whose weights $\hat{w}_i$ and $\hat{w}'_i$ are large.

	In our experiments, we minimize the objective function using the stochastic gradient descent method \citep{sutskever2013importance}. We discuss the computation time and convergence guarantees in \arefconv.
	
	From the penalty function \eqref{eq-penalty}, we can see why penalizing the upper bound on PIU guarantees individual-level fairness. As the penalty parameter value goes to infinity ($\lambda \rightarrow \infty$), the marginal probabilities $(\ePz,  \ePo)$ approach $(0, 0)$ or $(1, 1)$. This guarantees that the potential outcomes take the same value with probability $1$, which is sufficient to guarantee individual-level fairness. 

	\subsection{Comparison with existing fairness constraint} \label{subsec:feasible}
	
	To show the effectiveness of penalty function \eqref{eq-penalty}, we compare it with the constraint of the FIO method. 
	
	Suppose that our penalty function forces the upper bound on PIU to satisfy the following condition:
	\begin{align}
		\ePo (1 - \ePz) + (1 - \ePo) \ePz   \leq \delta. \label{eq5-p}
	\end{align} 
	Here we let constant $\delta$ be $\delta \in [0, 1]$ because otherwise we cannot force the PIU value to be less than $1$. 
	
	Meanwhile, the FIO constraint limits the mean unfair effect \eqref{eq-PE} to lie in 
	\begin{align}
	- \delta' \leq \ePo -  \ePz  \leq \delta' , \label{eq5-e}
	\end{align} 
	where $\delta' \in [0, 1]$ is a hyperparameter. If $\delta' = 0$, it ensures $\ePz$ $=$ $\ePo$.
	
	\begin{figure}[t]
		\includegraphics[height=3.6cm]{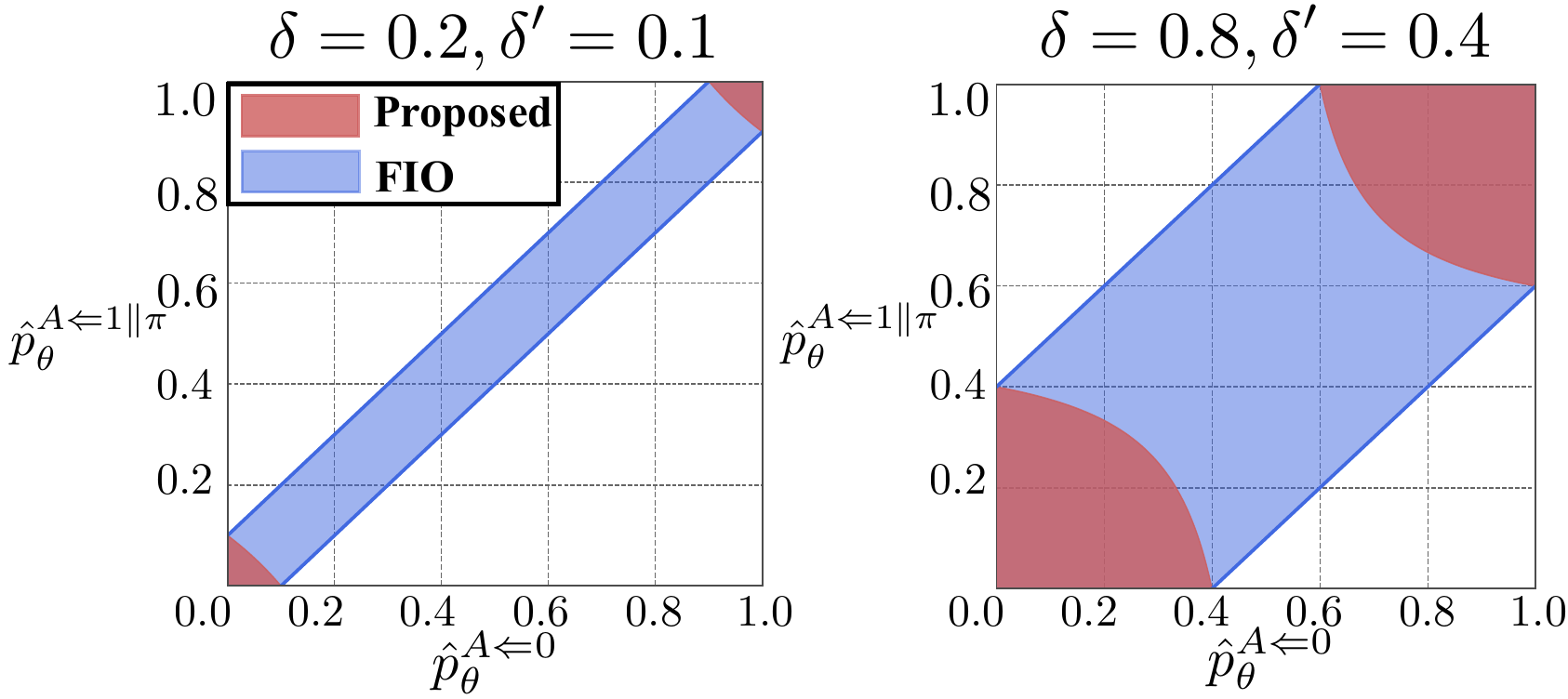}
		\centering 
		\caption{Feasible regions of our constraint (red) and FIO (blue) with ($\delta$, $\delta'$) $=$ (0.2, 0.1), (0.8, 0.4)} 
		\label{fig2-constrained}
	\end{figure}

	\Cref{fig2-constrained} shows the feasible region of our fairness condition (red) and the FIO constraint (blue), respectively, obtained by graphing the hyperbolic inequality in \eqref{eq5-p} and the linear inequality in \eqref{eq5-e}. Here, to clarify their difference, we consider the case where $\delta = 2 \delta'$.
	
	When $\delta \approx 0$, our fairness condition only accepts region $(\ePz, \ePo)$ $\approx$ $(0, 0)$ or $(1, 1)$, where the potential outcomes are likely to take the same value; hence, the unfair effect is likely to be zero. This demonstrates how effectively our condition removes an unfair effect for each individual. By contrast, with any $\delta'$ value, the FIO constraint always accepts point $(\ePz, \ePo)$ $=$ $(0.5, 0.5)$, where it is completely uncertain whether potential outcomes take the same value as detailed in \arefcomp. This implies that FIO predictions might be unfair for some individuals, indicating that FIO cannot ensure individual-level fairness.

	\subsection{Extension for dealing with latent confounders} \label{subsec:extension}
	
	So far, we have assumed that the marginal probabilities of potential outcomes can be estimated from data. This assumption, however, does not hold if there is a latent confounder \citep{pearl2009causality}, i.e., an unobserved variable that is a parent of the observed variables in the causal graph. Although this is possible in practice, inferring marginal probabilities becomes much more challenging.
	
	However, even in the presence of latent confounders, our method can ensure individual-level fairness if the lower and upper bounds on marginal probabilities are available. For instance, the bounds of \cite{miles2017partial} can be used when there is only a single {\it mediator} (i.e., a feature affected by sensitive feature $A$), and it takes discrete values. In such cases, we can achieve individual-level fairness by reformulating the penalty function as follows.
	
	Suppose that marginal probabilities $\pr(\Yz = 1)$ and $\pr(\Yo = 1)$ are bounded by
		\begin{align*}
		&\eLz \leq \pr(\Yz = 1) \leq \eUz  ,\\
		& \eLo \leq \pr(\Yo = 1) \leq \eUo ,
	\end{align*}	
	where $\eLz$, $\eUz$, $\eLo$, and $\eUo$ are the estimated lower and upper bounds, which we describe in  \arefextend. Then for any marginal probability values, the upper bound on PIU in \eqref{eq-Th1} is always smaller than twice the value of
	  	\begin{align}
	  	&G_{\theta}(\textbf{\textit{x}}_1, \dots \textbf{\textit{x}}_n) \nonumber \\
	  	&=  \eUo (1 - \eLz) + (1 - \eLo) \eUz .
	  	\label{eq-penalty_ex}
	  \end{align}	
	Therefore, by making this penalty function value nearly zero, we can achieve individual-level fairness. We experimentally confirmed that this framework makes fairer predictions than the original one in \arefextendedexp.

	\section{Experiments} \label{sec:experiment}
	
	We compared our method ({\bf Proposed}) with the following four baselines: (1) {\bf FIO} \citep{nabi2018fair}, (2) {\bf PSCF} \citep{chiappa2018path}, which aims to achieve individual-level fairness by assuming that the data are generated by additive noise models, (3) {\bf Unconstrained}, which does not use any constraints or penalty terms related to fairness, and (4) {\bf Remove} \citep[Section S4]{kusner2017counterfactual}, which removes unfair effects simply by making predictions without input features that correspond to the nodes on unfair pathways $\pi$. As classifiers of {\bf Proposed}, {\bf FIO}, {\bf Unconstrained}, and {\bf Remove}, we used a feed-forward neural network that contains two linear layers with $100$ and $50$ hidden neurons. Other settings are detailed in \arefexpset.
	
	{\bf Data and causal graphs}: For a performance evaluation, we used a synthetic dataset and two real-world datasets: the German credit dataset and the Adult dataset \citep{uci}. We sampled the synthetic data from the SEM, whose formulation is described in \arefsynthd. To define the unfair effect, we used the causal graph in \Cref{fig1}(b). With the real-world datasets, we evaluated the performance as follows. With the German dataset, we predicted whether each loan applicant is risky ($Y$) from their features such as gender $A$ and savings $\textbf{\textit{S}}$. With the Adult dataset, we predicted whether an annual income exceeds \$50,000 ($Y$) from features such as gender $A$ and marital status $M$. To measure unfairness, following \citep{chiappa2018path}, we used the causal graphs in \Cref{fig2}, which we detail in \arefreald. 
	
		\begin{table}[t] 
		\centering
		\caption{Test accuracy (\%) on each dataset}
		\label{table1}
		\scalebox{0.99}{
			\tabcolsep=0.8mm
			\begin{tabular}{lccc}
				\toprule
				Method & Synth & German & Adult \\		
				\midrule						
				{\bf Proposed}&     80.0 $\pm$ 0.9  & 75.0 &   75.2  \\
				{\bf FIO} &     	84.8 $\pm$ 0.6  & 78.0  &  81.2 \\
				{\bf PSCF} &     	74.8 $\pm$ 1.6  & 76.0  &  73.4 \\
				{\bf Unconstrained}  & 88.2 $\pm$ 0.9  & 81.0  &  83.2  \\
				{\bf Remove}  & 76.9 $\pm$ 1.3  & 73.0  &  74.7  \\
				\bottomrule
			\end{tabular}
		}
	\end{table}
	
	\begin{figure}[t]
		\includegraphics[height=2.17cm]{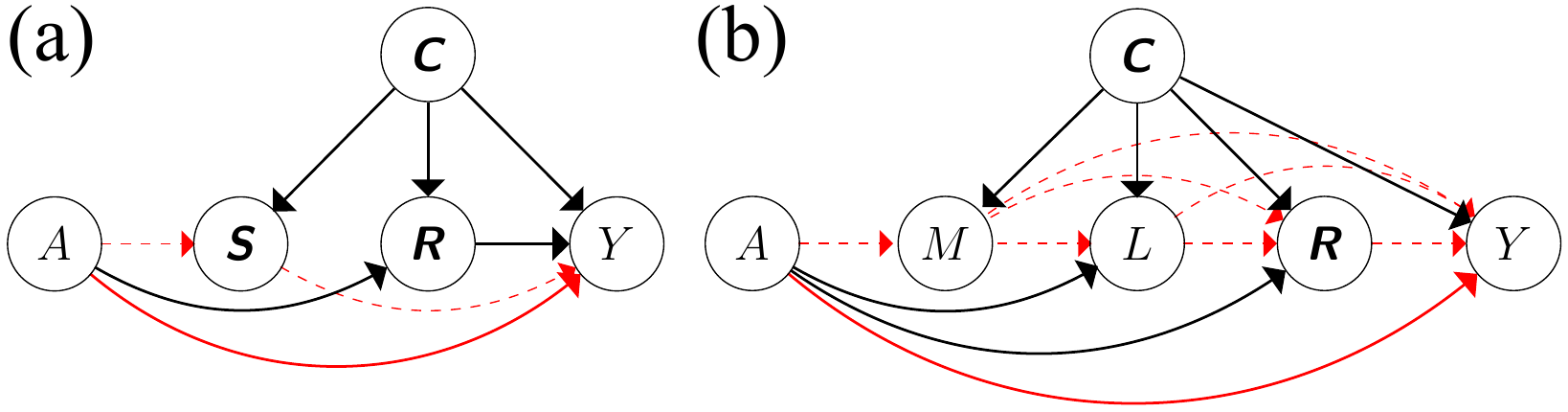}
		\centering 
		\caption{Causal graphs for (a) German credit dataset and (b) Adult dataset: Direct pathways with red solid edges are unfair. Red dashed pathway $A \rightarrow S \rightarrow Y$ is unfair in (a), and those that go through $M$ (i.e., $A$ $\rightarrow$ $M$ $\rightarrow$ $\cdots$ $\rightarrow$ $Y$) are unfair in (b).} 
		\label{fig2}
	\end{figure}

	{\bf Accuracy and fairness}: We evaluated the test accuracy and four statistics of the unfair effects: (\myra) the mean unfair effect \eqref{eq-PE}, (\myrb) the standard deviation in the conditional mean unfair effects in \eqref{eq-CE}, (\myrc) the upper bound on PIU, and (\myrd) the PIU.

	\Cref{table1} and \Cref{fig-ue} present the test accuracy and the four statistics of the unfair effects, respectively. In synthetic data experiments, we computed the means and the standard deviations based on $10$ experiments with randomly generated training and test data. \Cref{fig-ue} does not display two statistics (\myrb) and (\myrd) for the German and Adult datasets because computing them requires SEMs, which are unavailable for these real-world datasets. We do not show (\myra) or (\myrc) of \textbf{PSCF} because these are not well-defined for this method (see \arefsynthp\ for details).

	\begin{figure}[t]
		\includegraphics[height=6.4cm]{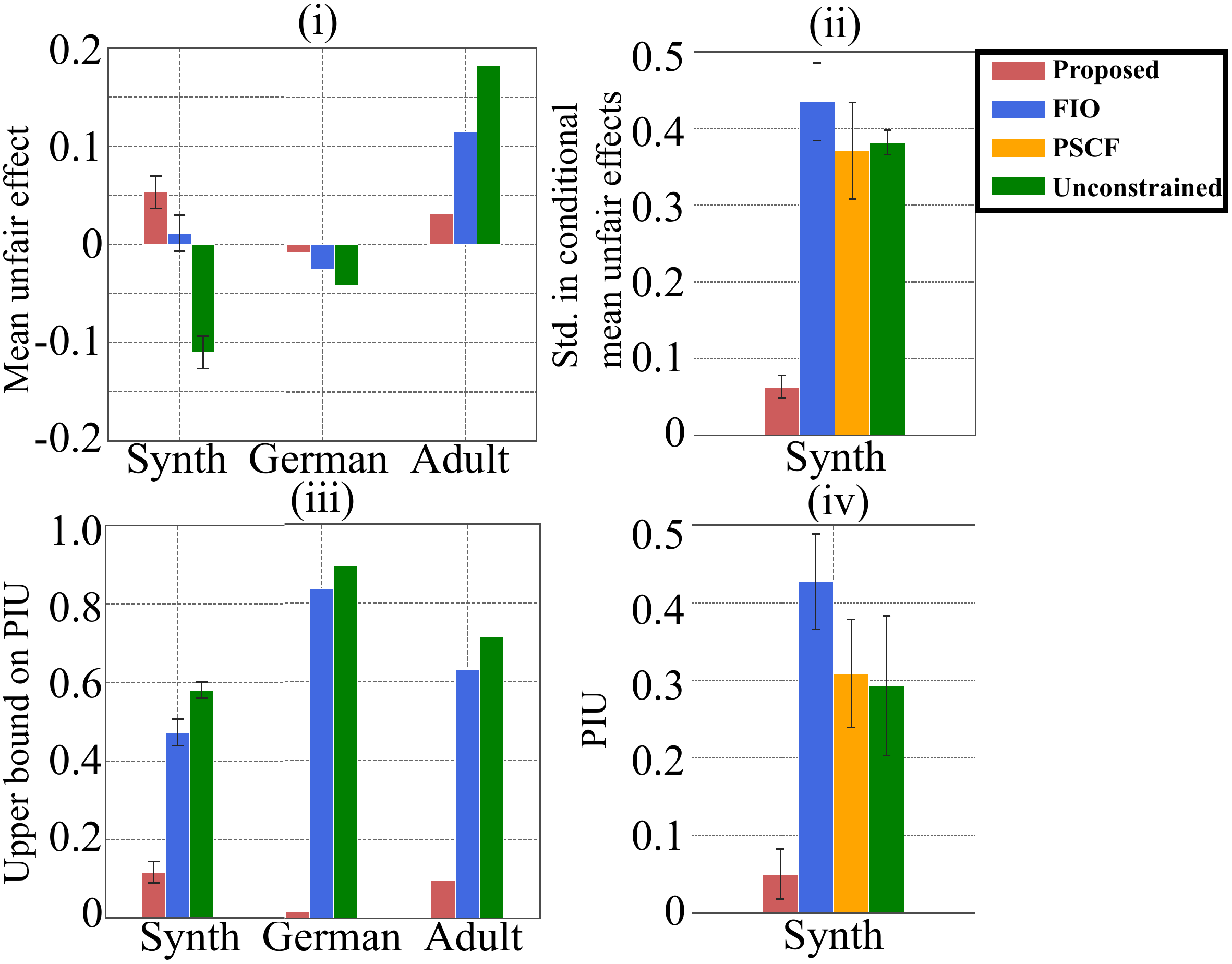}
	\centering 
	\caption{Four statistics of unfair effects on test data: The closer they are to zero, the fairer predictions are. Error bars of synthetic data (Synth) denote standard deviations in 10 runs with randomly generated data. With \textbf{Remove}, all statistics are zero (not shown). With {\bf PSCF}, (\myra) and (\myrc) are not well-defined as described in \arefsynthp. } 
	\label{fig-ue}
\end{figure}

	With {\bf Proposed}, all the statistics of the unfair effects were sufficiently close to zero, demonstrating that it made fair predictions for all individuals. This is because by imposing a penalty on the upper bound on PIU, {\bf Proposed} forced unfair effect values to be close to zero for all individuals, guaranteeing that the other statistics are close to zero.
	
	By contrast, regarding {\bf FIO} and {\bf PSCF}, the unfair effect values were much larger. With {\bf FIO}, although the mean unfair effect (i.e., (\myra)) was close to zero, the other statistics deviated from zero, indicating that constraining the mean unfair effect did not ensure individual-level fairness. {\bf PSCF} failed to reduce the value of the standard deviation in the conditional mean unfair effects (i.e., (\myrb)). This is because the data are not generated from additive noise models (see \arefsynthd\ for the data), violating the functional assumption of {\bf PSCF}. Since the large values of (\myrb) imply that unfair effects are greatly affected by the attributes of input features $\textbf{\textit{X}}$, these results indicate that {\bf FIO} and {\bf PSCF} made unfair predictions based on these attributes. With real-world datasets, {\bf FIO} provided large values of the upper bound on PIU (i.e., (\myrc)). These upper bound values cast much doubt on whether the predictions of {\bf FIO} are fair for each individual, which is problematic in practice.
	
	The test accuracy of {\bf Proposed} was lower than {\bf FIO}, higher than {\bf Remove}, and comparable to {\bf PSCF}. Since {\bf FIO} imposes a much weaker fairness constraint than {\bf Proposed}, {\bf Remove}, and {\bf PSCF} (i.e., the methods designed for achieving individual-level fairness), it achieved higher accuracy. By contrast, since {\bf Remove} eliminates all informative input features that are affected by the sensitive feature to guarantee individual-level fairness, it provided the lowest accuracy. The comparison of {\bf Proposed} and {\bf PSCF} indicates that although our method employs a more severe fairness condition than {\bf PSCF}, it barely sacrifices prediction accuracy, demonstrating that it strikes a better balance between individual-level fairness and accuracy.
	
	\begin{figure}[t]
		\includegraphics[height=6.45cm]{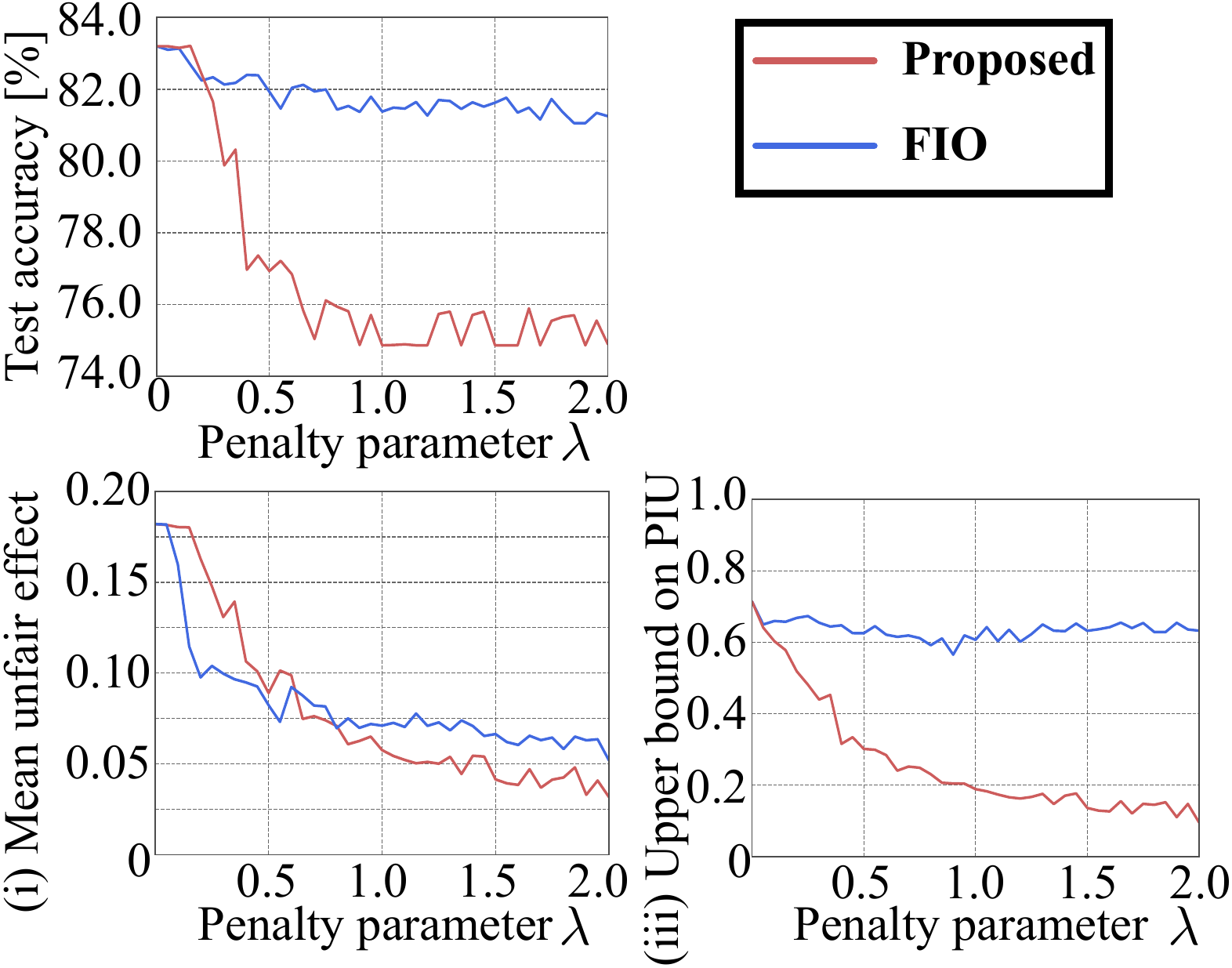}
		\centering 
			\caption{Test accuracy, mean unfair effect, and upper bound on PIU of (a) {\bf Proposed} (red) and (b) {\bf FIO} (blue) for Adult dataset when increasing penalty parameter value as $\lambda$ $=$ $0$, $0.05$, $0.10$, $\cdots$, $2.00$.} 
		\label{fig3}
	\end{figure}
	
	{\bf Penalty parameter effects}: Using the Adult dataset, we further compared the performance of {\bf Proposed} with {\bf FIO} using various penalty parameter values. Regarding unfair effects, we evaluated two statistics (\myra) the mean unfair effect and (\myrc) the upper bound on PIU since the others are unavailable with this real-world dataset, as already described above.
	
	 \Cref{fig3} presents the results. As the penalty parameter value increased, the test accuracy of {\bf Proposed} decreased more sharply than {\bf FIO} since {\bf Proposed} imposed a stronger fairness condition. However, with {\bf Proposed}, both (\myra) and (\myrc) dropped to nearly zero, while only (\myra) decreased with {\bf FIO}. This implies that while it remains uncertain whether the predictions of {\bf FIO} are individually fair, the predictions of {\bf Proposed} are more reliable since it can successfully reduce the upper bound on PIU (i.e., (\myrc)) and guarantee individual-level fairness, which is helpful for practitioners.
	
	{\bf Additional experimental results}: To further demonstrate the effectiveness of our method, we present several additional experimental results in \arefaddexp. Through our experiments, we show that our method also worked well using other classifier than the neural network (\areflogisticexp), present the statistical significance of the test accuracy (\areferrbexp), confirm that both our method and {\bf PSCF} achieved individual-level fairness when the data satisfy the functional assumptions (\arefadmexp), demonstrate the tightness of our upper bound on PIU (\arefboundexp), and evaluate the performance of our extended framework for addressing latent confounders (\arefextendedexp).

	\section{Related work} \label{sec:related}
	
	{\bf Causality-based fairness}: Motivated by recent developments in inferring causal graphs \citep{chikahara2018causal,glymour2019review}, many causality-based approaches to fairness have been proposed \citep{chiappa2018path,kilbertus2017avoiding,kusner2017counterfactual,kusner2019making,nabi2018fair,nabi2019learning,russell2017worlds,salimi2019interventional,wu2018discrimination,wu2019counterfactual,xu2019achieving,zhang2016causal,zhang2017anti,zhang2018causal,zhang2018equality,zhang2018fairness}. However, few are designed for making individually fair predictions due to the difficulty of estimating the conditional mean unfair effect in \eqref{eq-CE}. Although {\it path-specific counterfactual fairness} (PC-fairness) \citep{wu2019pc} was proposed to estimate its lower and upper bounds, it only addresses measuring unfairness in data and not making fair predictions.
	
	By contrast, we established a learning framework for making individually fair predictions under much weaker assumptions than the existing approaches.

	{\bf Bounding PIU}: To learn an individually fair classifier, our framework utilizes the upper bound on PIU in \eqref{eq-Th1}. Compared with the existing bounds described below, this upper bound has the following two advantages.
	
	First, it can be used for binary potential outcomes. If we consider continuous potential outcomes, we can use several bounds on a functional of the joint distribution of potential outcomes \citep{fan2017partial,firpo2019partial} because PIU is also such a functional. However, these existing bounds cannot be used in our binary classification setting where PIU is formulated based on binary potential outcomes.
	
	Second, it is much tighter than the result in \cite{rubinstein2017combinatorial}, which provides an upper bound on an expectation of random variables whose joint distribution is unavailable.\footnote{Such an upper bound is known as {\it correlation gap} in the field of robust optimization. \citep{agrawal2010correlation}.} Since PIU can be written as an expectation (i.e., $\E_{\Yz, \Yo}[\I(\Yz \neq \Yo)]$), we can apply this result to PIU; however, the bound becomes much looser than ours. Although the multiplicative constant in \eqref{eq-Th1} is $2$, this value becomes $200$ with the bound of \cite{rubinstein2017combinatorial}. If we use such a loose upper bound, we need to impose an excessively severe penalty on it to ensure that PIU is close to zero. By contrast, using our upper bound, we can avoid imposing such a severe penalty, thus preventing an unnecessary decrease in prediction accuracy.

	\section{Conclusion}
	
	We proposed a learning framework for guaranteeing individual-level fairness without impractical functional assumptions. Based on the concept called path-specific causal effects, we defined PIU and derived its upper bound that can be estimated from data. By forcing this upper bound value to be nearly zero, our proposed method trains an individually fair classifier. We experimentally show that our method makes individually fairer predictions than the existing methods at a slight cost of accuracy, indicating that it strikes a better balance between fairness and accuracy.

	\bibliographystyle{plainnat}
	\bibliography{main.bib}

	\onecolumn
	\appendix
	\setcounter{equation}{0}
	\setcounter{figure}{0}
	\renewcommand{\theequation}{A\arabic{equation}}
	\renewcommand{\thefigure}{A\arabic{figure}}

	\clearpage
	
	\begin{center}
		{\fontsize{18pt}{0pt}\selectfont \bf Appendix}
	\end{center}

	\section{Path-specific causal effects} \label{sec-a:pod}
	
	In this section, we provide a formal definition of path-specific causal effects \citep{avin2005identifiability}. We begin by reviewing a formulation of path-specific causal effects in \Cref{subsec-a:rpo} and then formally define SEM and the concept of {\it intervention} \citep{pearl2009causality} in \Cref{subsec-a:dcm}, both of which are needed to define path-specific causal effects. Finally, we define path-specific causal effects in \Cref{subsec-a:pse}.

\subsection{Revisiting example}  \label{subsec-a:rpo}
	
	To take an example of path-specific causal effects, we revisit a scenario of hiring decisions for physically demanding jobs with the causal graph in \Cref{fig1}(b). 
	
	In this scenario, as described in \Cref{subsec:pse}, the path-specific causal effect on prediction $Y$ is formulated as the difference $\Yo - \Yz$, where $\Yz$ and $\Yo$ are the following potential outcomes:
		\begin{align}
		\tag{\ref{po}}
		\begin{aligned}
			\Yz = h_{\theta}(0, Q, D(0), M(0)), \quad \Yo = h_{\theta}(1, Q, D(1), M(0)).
		\end{aligned}
	\end{align}
	Here $D(a)$ and $M(a)$ ($a \in \{0, 1\}$) represent the (counterfactual) attributes of number of children $D$ and physical strength $M$. Using structural equations \eqref{causalmodel1}, these attributes are formulated as
		\begin{align}
		D(a) = a + U_D Q, \quad M(a) = 3a + 0.5 Q + U_M. 	\tag{\ref{pm}}
	\end{align}
	
	To illustrate these attributes, suppose that a woman $i \in \{1, \dots, n\}$ has the attributes $\textbf{\textit{x}}_i = \{A=0, Q=q_i, D=d_i, M=m_i\}$. For her, attributes $D(0)$ and $M(0)$ are observed as $d_i$ and $m_i$, respectively; that is,
	\begin{align*}
		D(0) =d_i =  u_{D, i}  q_i, \quad M(0) = m_i = 0.5q_i + u_{M, i}
	\end{align*}
	where $u_{D, i}$ and $u_{M, i}$ are the values of unobserved noises $U_D$ and $U_M$ for woman $i$, respectively. Meanwhile, $D(1)$ and $M(1)$ express the counterfactual attributes of $D$ and $M$ that would be if she were male, respectively, which are formulated as
	\begin{align*}
		D(1) = 1 + u_{D, i}  q_i, \quad M(1) = 3 + 0.5 q_i + u_{M, i}.
	\end{align*}
Therefore, for her, potential outcome $\Yz$ in \eqref{po} is provided by directly using her attributes $A=0$, $Q=q_i$, $D(0)=d_i$, and $M(0)=m_i$. Meanwhile, $\Yo$ is obtained using her attributes $Q=q_i$ and $M(0)=m_i$ and counterfactual attributes $A=1$ and $D(1)=1 + u_{D, i} q_i$. 
	
	To take another example, suppose that a man $j \in \{1, \dots, n\}$ $(j \neq i)$ has the attributes $\textbf{\textit{x}}_j = \{A=1, Q=q_j, D=d_j, M=m_j\}$. For him, attributes $D(1)$ and $M(1)$ are observed as $d_j$ and $m_j$, respectively; in other words,
		\begin{align*}
		D(1) = d_j = 1 + u_{D, j} q_j,\quad M(1) = m_j = 3 + 0.5 q_j + u_{M, j}.
	\end{align*}
	In contrast, $D(0)$ and $M(0)$ represent the counterfactual attributes of $D$ and $M$ that would be if he were female, respectively, which are expressed as
		\begin{align*}
		D(0) = u_{D, j} q_j. \quad M(0) = m_j = 0.5q_j + u_{M, j}.
	\end{align*}
	Consequently, for this man, potential outcome $\Yz$ in \eqref{po} is provided by using his attribute $Q=q_j$ and counterfactual attributes $A=0$, $D(0)=u_{D, j} q_j$, and $M(0)=0.5q_j + u_{M, j}$. By contrast, $\Yo$ is obtained by using his attributes $A=1$, $Q=q_j$, $D(1)=d_j$ and counterfactual attribute $M(0)=0.5q_j + u_{M, j}$.
	
	Without an SEM, since unobserved noise values (i.e., $u_{D,i}$, $u_{D, j}$, $u_{M, i}$ and $u_{M, j}$) are unavailable, we cannot obtain counterfactual attributes $D(1)$ and $M(1)$ for woman $i$ and $D(0)$ and $M(0)$ for man $j$. However, if the SEM is available, we can compute them by sampling these noises from their distribution $\pr(\textbf{\textit{U}})$ and computing the counterfactual attributes based on \eqref{pm}. In the next section, we formally define an SEM and formulate these counterfactual attributes. 
	 
	\subsection{Concepts of causal inference} \label{subsec-a:dcm}
	
	\subsubsection{SEM} \label{subsubsec-a:sem}
	
	SEM $\mathcal{M}$ is formally defined as quadruplet $\mathcal{M} = (\textbf{\textit{U}}, \textbf{\textit{V}}, \textbf{\textit{F}}, \pr(\textbf{\textit{U}}))$, where $\textbf{\textit{U}}$ is a set of unobserved noise variables called {\it exogenous variables}, $\textbf{\textit{V}}$ is a set of observed variables called {\it endogenous variables}, which are represented in a causal graph, $\textbf{\textit{F}}$ is a set of deterministic functions, and $\pr(\textbf{\textit{U}})$ is the joint distribution over $\textbf{\textit{U}}$ \citep{pearl2009causality}.
	
	For each variable $V \in \textbf{\textit{V}}$, the SEM describes how its variable value is determined, formulated as a structural equation:
	\begin{align}
		V = f_V(\textbf{\textit{pa}}(V), \textbf{\textit{U}}_V), \label{def_se}
	\end{align}
	where $f_V \in \textbf{\textit{F}}$ is a deterministic function, $\textbf{\textit{pa}}(V) \subseteq \textbf{\textit{V}} \backslash V$ are the variables that are the parents of $V$ in the causal graph, and $ \textbf{\textit{U}}_V \subseteq \textbf{\textit{U}}$.

	In our setting, we consider SEM $\mathcal{M}^{p}$, whose endogenous variables $\textbf{\textit{V}} = \{\textbf{\textit{X}}, Y\}$ contain prediction $Y$. In this SEM, each input feature variable $V \in \textbf{\textit{X}}$ is expressed by structural equation \eqref{def_se}. By contrast, as described in \Cref{subsec:pse}, the structural equation over prediction $Y$ is expressed by classifier $h_{\theta}$. If it is deterministic, the structural equation is given by
	\begin{align*}
		Y = h_{\theta}(\textbf{\textit{X}}),
	\end{align*}
	and if $h_{\theta}$ is a probabilistic classifier, it is expressed by
	\begin{align*}
		Y = h_{\theta}(\textbf{\textit{X}}, \textbf{\textit{U}}_Y ),
	\end{align*}
	where $\textbf{\textit{U}}_Y \subseteq \textbf{\textit{U}}$ denotes the unobserved random noises used in the classifier. These formulations of the structural equation over prediction $Y$ can be regarded as a special case of $\eqref{def_se}$, where deterministic function $f_Y \in \textbf{\textit{F}}$ is replaced with classifier $h_{\theta}$.
	
	Note that this SEM differs from an SEM in the standard setting of causal inference, which expresses a true generating process of the observed data. While the former SEM includes prediction $Y$, the latter contains observed decision outcome $Y$, whose observations are included in the data and represented by a structural equation \eqref{def_se}.  

	\subsubsection{Interventions} \label{subsec-a:int}
	
	To define potential outcomes $\Yz$ and $\Yo$, we need to formulate the counterfactual attributes of each individual that would be if sensitive feature attribute were changed (e.g., $D(a)$ and $M(a)$ in \eqref{pm}), which requires an operation for SEM $\mathcal{M}^p$, called an intervention \citep{pearl2009causality}. 
	
	To express the counterfactual situations where sensitive feature attribute is changed, we consider intervention $do(A=a)$ on sensitive feature $A$, which forces $A$ to take a certain value, $a \in \{0, 1\}$. This intervention is formally defined as a replacement of the structural equation over $A$ with $A = a$. 
	
	Such a replacement modifies the structural equations over the descendant variables of $A$, and the counterfactual attributes can be formulated by this modified SEM. For instance, when the causal graph in \Cref{fig1}(b) is given, intervention $do(A=a)$ modifies the structural equations over $D$ and $M$ as \eqref{pm}, which provides counterfactual attributes $D(a)$ and $M(a)$.
	
	In general, an SEM modified by an intervention is called an {\it interventional SEM}. In what follows, let $\mathcal{M}^p_{A=a}$ denote an interventional SEM that is obtained by performing intervention $do(A=a)$.
	
	\subsection{Defining path-specific causal effects} \label{subsec-a:pse}
	
	Using interventional SEM $\mathcal{M}^p_{A=a}$ ($a \in \{0, 1\}$) defined in \Cref{subsec-a:int}, we define path-specific causal effects (see the original paper \citep{avin2005identifiability} for details). As already described in \Cref{subsec:pse}, a path-specific causal effect is formulated by the difference between two potential outcomes, $\Yz$ and $\Yo$, as $\Yo - \Yz$. In what follows, we formally define these potential outcomes.
	
	{\bf Definition}: To define potential outcome $\Yz$, we consider interventional SEM $\mathcal{M}^p_{A=0}$, which is obtained by simply performing intervention $do(A=0)$. Suppose that this SEM expresses each variable $V \in \{\textbf{\textit{X}}, Y\}$ by the following structural equation:
	\begin{align}
		V = f_V (\textbf{\textit{pa}}(V)_{A=0}, \textbf{\textit{U}}_V), \label{modifiedSE_0}
	\end{align}
	where $\textbf{\textit{pa}}(V)_{A=0}$ denotes variables $\textbf{\textit{pa}}(V)$ (i.e., parents of variable $V$), whose values are determined by interventional SEM $\mathcal{M}^p_{A=0}$. Then potential outcome $\Yz$ is defined as prediction $Y$, whose structural equation is expressed by \eqref{modifiedSE_0} where function $f_Y$ is given by classifier $h_{\theta}$.

	By contrast, to define potential outcome $\Yo$, we need an SEM that is modified using interventional SEMs $\mathcal{M}^p_{A=0}$ and $\mathcal{M}^p_{A=1}$. To formulate this modified SEM, for each variable $V \in \{\textbf{\textit{X}}, Y\}$, we partition its parents $\textbf{\textit{pa}}(V)$ into two subsets, $\textbf{\textit{pa}}(V) = \{\textbf{\textit{pa}}(V)^{\pi}, \textbf{\textit{pa}}(V)^{\overline{\pi}}\}$, where $\textbf{\textit{pa}}(V)^{\pi}$ is the members of  $\textbf{\textit{pa}}(V)$ connected with $V$ on unfair pathways $\pi$, and $\textbf{\textit{pa}}(V)^{\overline{\pi}}$ is a complementary set (i.e., $\textbf{\textit{pa}}(V)^{\overline{\pi}} = \textbf{\textit{pa}}(V) \backslash \textbf{\textit{pa}}(V)^{\pi}$). Based on these two subsets, we consider the following structural equation over $V  \in \{\textbf{\textit{X}}, Y\}$:
	\begin{align}
		V = f_V (\textbf{\textit{pa}}(V)^{\pi}_{A=1}, \textbf{\textit{pa}}(V)^{\overline{\pi}}_{A=0}, \textbf{\textit{U}}_V), \label{modifiedSE}
	\end{align}
	where $\textbf{\textit{pa}}(V)^{\pi}_{A=1}$ is a set of the variables in $\textbf{\textit{pa}}(V)^{\pi}$ whose values are determined by interventional model $\mathcal{M}^p_{A=1}$, and $\textbf{\textit{pa}}(V)^{\overline{\pi}}_{A=0}$ is a set of the variables in $\textbf{\textit{pa}}(V)^{\overline{\pi}}$ whose values are provided by $\mathcal{M}^p_{A=0}$. Then potential outcome $\Yo$ is defined as prediction $Y$, whose structural equation is represented by \eqref{modifiedSE}.

	In this way, $\Yo$ is formulated by performing intervention $do(A=1)$ only on the variables that involve unfair pathways $\pi$ (i.e., $\textbf{\textit{pa}}(V)^{\pi}$), and by taking the difference between $\Yz$ and $\Yo$, we can measure the influence via the pathways $\pi$. 
	
	
	 {\bf Example}: We provide a simple formulation example of potential outcomes $\Yz$ and $\Yo$ based on the causal graph in \Cref{fig1} (a).
	 
	 Since the parents of prediction $Y$ in this causal graph are $\textbf{\textit{pa}}(Y)=\textbf{\textit{X}}=\{A, Q, D, M\}$, by simply performing intervention $do(A=0)$ on these variables, potential outcome $\Yz$ is formulated as 
	\begin{align}
		\Yz = h_{\theta}(0, Q, D(0), M(0), \textbf{\textit{U}}_Y), \label{pyz1}
		\end{align}
	where classifier inputs $\{0, Q, D(0), M(0)\}$ correspond to $\textbf{\textit{pa}}(Y)_{A=0}$, whose values are given by interventional SEM $\mathcal{M}^p_{A=0}$.
	
	
	Potential outcome $\Yo$ is defined based on unfair pathway $\pi = \{A \rightarrow Y\}$. Since prediction $Y$ is not connected with $Q$, $D$ or $M$ on unfair pathway $\pi$ but connected with $A$, the parents of prediction $Y$ (i.e., $\textbf{\textit{pa}}(Y)=\textbf{\textit{X}}$) is partitioned into two subsets, $\textbf{\textit{pa}}(Y)^{\overline{\pi}} = \{Q, D, M\}$ and $\textbf{\textit{pa}}(Y)^{\pi} = \{A\}$. Letting the values of these variable subsets be determined by interventional models $\mathcal{M}^p_{A=0}$ and $\mathcal{M}^p_{A=1}$, respectively, potential outcome $\Yo$ is expressed as
	\begin{align}
		\Yo = h_{\theta}(1, Q, D(0), M(0), \textbf{\textit{U}}_Y), \label{pyo1}
	\end{align}
	where classifier inputs $\{Q, D(0), M(0)\}$ and $\{1\}$ correspond to $\textbf{\textit{pa}}(Y)^{\overline{\pi}}_{A=0}$ and $\textbf{\textit{pa}}(Y)^{\pi}_{A=1}$, respectively. 

	Given two potential outcomes $\Yz$ and $\Yo$, by taking their difference (i.e., $\Yo - \Yz$), we can measure the influence via direct pathway $\pi = \{A \rightarrow Y\}$, which is called {\it natural direct causal effects} \citep{pearl2001direct}. 
	
	Note that we can completely remove this direct causal effect by making a prediction without sensitive feature $A$ \citep[Section S4]{kusner2017counterfactual}. However, if we consider multiple unfair pathways, we need to remove all the features that are descendants of $A$ in the causal graph, which may seriously decrease the prediction accuracy, as described in \Cref{subsec:problem}.
	
	\section{Assumptions} \label{sec-a:asmp}
	
	To estimate the marginal probabilities of potential outcomes $\pr(\Yz = 1)$ and $\pr(\Yo = 1)$, our method uses two standard assumptions, both of which are widely used in the existing methods \citep{chiappa2018path,nabi2018fair,zhang2017anti,zhang2016causal}.
	
	One is an assumption on unfair pathways $\pi$, which is expressed using the following graphical condition called the {\it recanting witness criterion}:
	 \begin{definition}[Recanting witness criterion \citep{avin2005identifiability}]\label{recanting}
		Given pathways $\pi$, let $Z$ be a node in the causal graph that satisfies the following:
			\setlength{\leftmargini}{15pt}
		\begin{enumerate}
			\item There is a pathway from $A$ to $Z$ ($A \rightarrow \dots \rightarrow Z$) in $\pi$.
			  \item	There is a pathway from $Z$ to $Y$ ($Z \rightarrow \dots \rightarrow Y$) in $\pi$.
			  \item There is another pathway from $Z$ to $Y$ ($Z \rightarrow \dots \rightarrow Y$) that is in the causal graph but not in $\pi$.
			\end{enumerate}
		Then pathways $\pi$ satisfy the recanting witness criterion with node $Z$, which is called a witness.
		\end{definition}
	For example, consider the causal graph in \Cref{kite_bow}(a), where the unfair pathway is $\pi = \{A \rightarrow M_1 \rightarrow M_2 \rightarrow Y\}$. Clearly, pathway $\pi$ satisfies the recanting witness criterion with witness $M_1$. 
	
	\begin{figure}[t]
		\includegraphics[width=11.5cm]{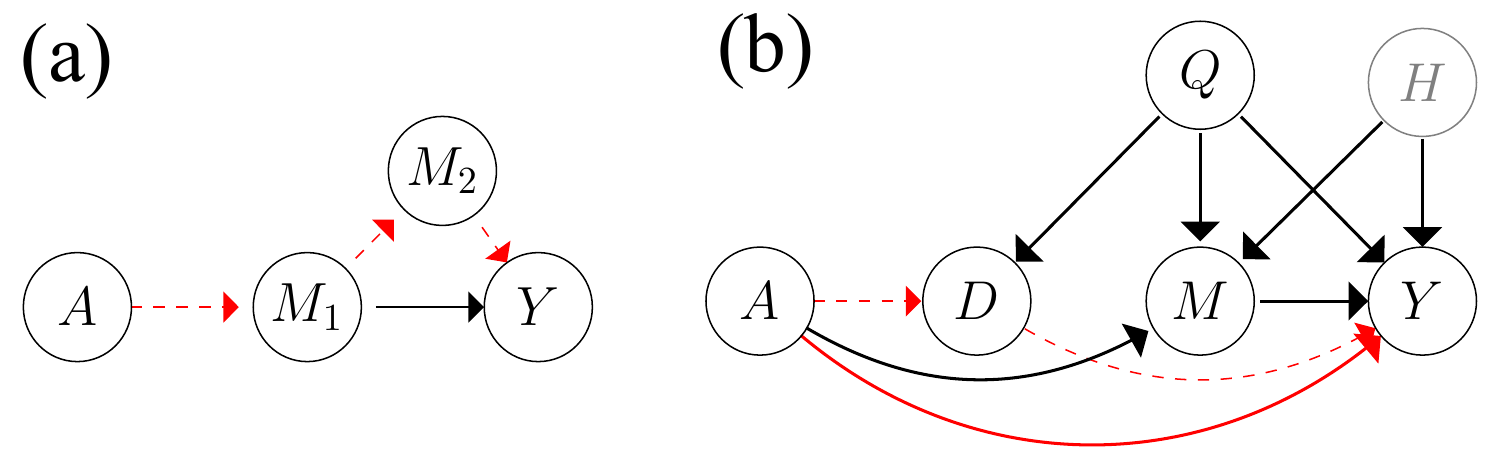}
		\centering 
		\caption{Two causal graphs that violate \Cref{recanting_asmp,ignorability}, respectively. Unfair pathways are (a): $\pi = \{A \rightarrow M_1 \rightarrow M_2 \rightarrow Y\}$ and (b): $\pi = \{A \rightarrow Y, A \rightarrow D \rightarrow Y\}$.}
		\label{kite_bow}
	\end{figure}

	According to \citet{avin2005identifiability}, to estimate marginal probabilities $\pr(\Yz = 1)$ and $\pr(\Yo = 1)$, we need to assume that pathways $\pi$ do {\bf not} satisfy the recanting witness criterion in \Cref{recanting}; that is,
	\begin{assumption}\label{recanting_asmp}
		Pathways $\pi$ do {\bf not} satisfy the recanting witness condition.		
	\end{assumption}

	The other is a common assumption in causal inference called {\it conditional ignorability} \citep{rosenbaum1983central}, which requires conditional independence relations between variables. 
	
	We formulate this assumption based on the estimators in a previous work \citep{huber2014identifying}, which we use in our method as presented in \eqref{mp_huber}.  As an example, we show a formulation based on the causal graph in \Cref{fig1}(b). 
	
	In what follows, we use the same notations as those in the original paper \citep{huber2014identifying}. Let potential outcomes $\Yz$ and $\Yo$ denote
	\begin{align}
		\label{notat_huber}
		\begin{aligned}
		\Yz = Y(0, D(0), M(0)), \quad \Yo = Y(1, D(1), M(0)),
		\end{aligned}
		\end{align}
	respectively, where $D(0)$, $D(1)$, and $M(0)$ express counterfactual attributes formulated by \eqref{pm}. Then the conditional ignorability is expressed as follows:
	 \begin{assumption}[Conditional ignorability \citep{huber2014identifying}]\label{ignorability}
	 For all $a, a', a'' \in \{0, 1\}$ and $d, m, q$ in the supports of $D$, $M$, and $Q$, the following four relations hold:
	 \begin{align}
	 		&\{Y(a, d, m), D(a'), M(a'')\} \indep A | Q = q	 \label{ig1}	\\
	 		&Y(a', d, m) \indep D | A = a, Q = q  \label{ig2}\\
	 		&Y(a', d, m) \indep M | A = a, Q = q \label{ig3}\\
	 		& \pr(A=a | Q=q, D=d, M=m) > 0.	 \label{ig4}
	 	\end{align}	 
	 \end{assumption}
	
	In \Cref{ignorability}, three relations, \eqref{ig1}, \eqref{ig2}, and \eqref{ig3}, are needed to express the potential outcome using the observed data distribution. As mentioned by \citet{huber2014identifying}, these relations are not satisfied if there is an unobserved variable called a latent confounder, which is an unobserved parent of the observed variables. For instance, when the causal graph in \Cref{kite_bow}(b) is given, the aforementioned relations do not hold due to a latent confounder, $H$ (gray node). However, even in the presence of latent confounders, in some cases, our method can achieve individual-level fairness using an extended penalty function, as described in \Cref{subsec:extension}.

	The last relation \eqref{ig4} is used to avoid a division by zero when computing the estimators \eqref{mp_huber}.

	\section{Proof of Theorem 1} \label{sec-a:th1}

	We first introduce several notations. Let the marginal potential outcome probabilities that satisfy $\Yz=1$ and $\Yo=1$ be $\mz$ and $\mo$, and their joint probabilities, $(\Yz, \Yo) = (0, 0)$, $(0, 1)$, $(1, 0)$, and $(1, 1)$, be $\pzz$, $\pzo$, $\poz$, and $\poo$. 
	
	Then we have
	\begin{align}
	\begin{aligned}
	\poz + \poo = \mz, \quad \pzz + \pzo = 1 - \mz,\quad \pzo + \poo = \mo,\quad  \text{and} \quad \poz + \pzz = 1 - \mo.
	\end{aligned}
	\label{eq:marginal}
	\end{align}
	
	As described in \Cref{subsubsec:ub}, the right-hand side in \eqref{eq-Th1} in \Cref{th1} can be represented by marginal probabilities. With the above notations, it can be written as $2(\mo(1-\mz) + \mz(1-\mo))$.
	
	Therefore, our goal is to prove
	\[
	\pzo + \poz \le 2(\mo(1-\mz) + \mz(1-\mo)).
	\]
	Since all the joint probabilities in \eqref{eq:marginal} are non-negative, $\pzo$ and $\poz$ become at most $\min\{\mo$, $1 - \mz \}$ and $\min\{\mz$, $1 - \mo\}$, respectively, yielding:
	\begin{align}
	\pzo + \poz \le \min\{\mo, 1 - \mz \} + \min\{\mz, 1 - \mo\}. \label{piu-ub}
	\end{align}
	Hence, it suffices to prove
	\begin{equation}\label{eq:target}
	\min\{\mo, 1 - \mz \} + \min\{\mz, 1 - \mo\} \le 2\mo(1-\mz) + 2\mz(1-\mo).
	\end{equation}
	
		\begin{figure}[t]
		\includegraphics[width=4cm]{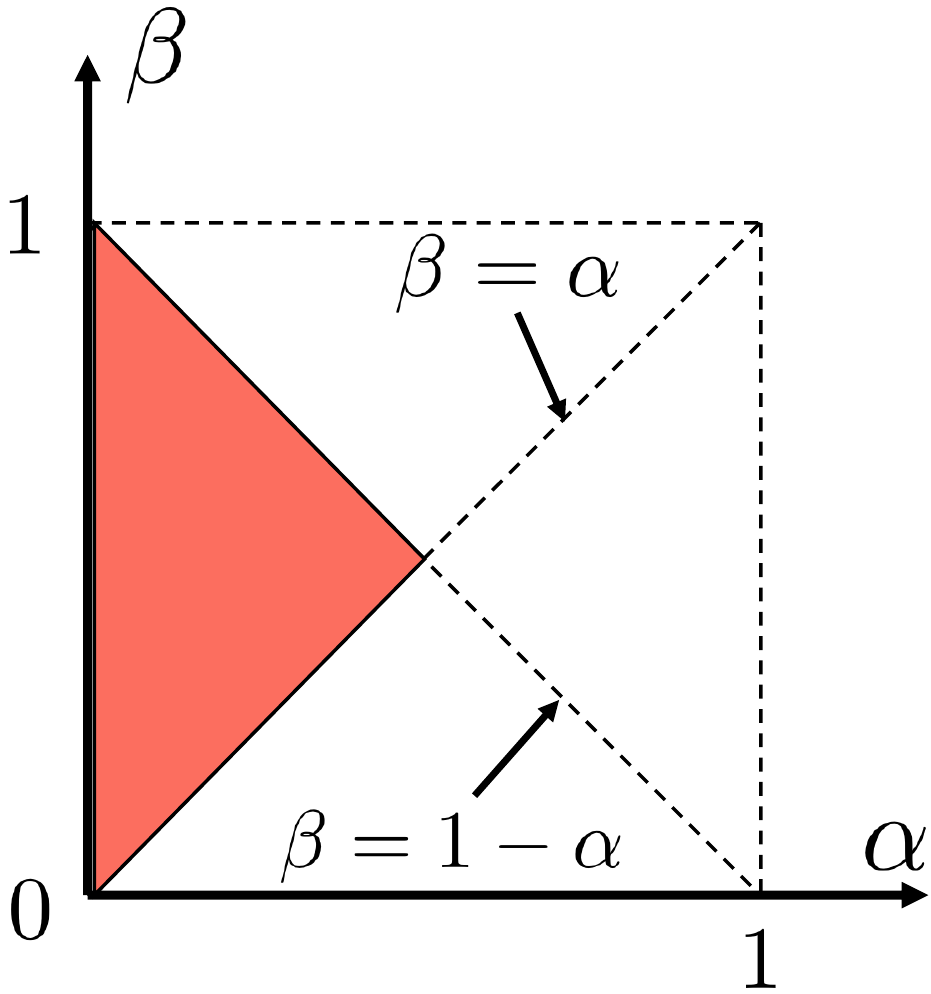}
		\centering 
		\caption{Red area represents region where marginal probability values $\alpha$ and $\beta$ satisfy $\mz \le \mo \le 1-\mz$}
		\label{fig-proof}
	\end{figure}
	
	Since both sides in \eqref{eq:target} are symmetrical with respect to $\mo = \mz$ and $\mo =1 - \mz$, it is sufficient to consider the case when $\mz \le \mo \le 1-\mz$, which is illustrated in \Cref{fig-proof} as the red region.

	In this case, since $\min\{\mo$, $1 - \mz \}$ $=$ $\mo$ and $\min\{\mz, 1 - \mo\}$ $=$ $\mz$, \eqref{eq:target} is reduced to 
	\begin{align}
	\mo + \mz &\le 2\mo(1-\mz) + 2\mz(1-\mo) \quad \mz + \mo -4\mz\mo \ge 0 .\label{eq-sup}
	\end{align}
	Since $\mz + \mo \le 1$ holds in this case, we have inequality $\mz + \mo - (\mz + \mo)^2 \ge 0$. Using this inequality, inequality \eqref{eq-sup} can be proven as follows:
	\begin{align}
	\mz + \mo -4\mz\mo &= \mz + \mo - (\mz + \mo)^2 + (\mz - \mo)^2 \ge 0.
	\end{align}
	Thus, we prove \Cref{th1}.

		\section{Derivation of (9)} \label{sec-a:mp_huber}
		
		Following the original paper \citep{huber2014identifying}, we derived the following formulation of the existing estimators of marginal potential outcome probabilities:
			\begin{align}
			\begin{aligned}
				\ePz = \frac{1}{n} \sum_{i=1}^n \I(a_i = 0) \hat{w}_i c_{\theta}(a_i, q_i, d_i, m_i), \quad \ePo = \frac{1}{n} \sum_{i=1}^n \I(a_i = 1) \hat{w}_i' c_{\theta}(a_i, q_i, d_i, m_i),
			\end{aligned}
			\tag{\ref{mp_huber}}
		\end{align}
		where $c_{\theta}(\textbf{\textit{X}}) = \pr(Y=1 | \textbf{\textit{X}})$ is the conditional distribution given by classifier $h_{\theta}$, $\I(\cdot)$ is an indicator function, and $\hat{w}_i$ and $\hat{w}_i'$ are the following weights: 
		\begin{align*}
			\hat{w}_i = \frac{1}{\hat{\pr}(A=0 | q_i)}, \quad \hat{w}_i' = \frac{\hat{\pr}(A=1 | q_i, d_i) \hat{\pr}(A=0 | q_i, d_i, m_i)}{\hat{\pr}(A=1 | q_i) \hat{\pr}(A=0 | q_i, d_i) \hat{\pr}(A=1 | q_i, d_i, m_i)},
		\end{align*}
		 where $\hat{\pr}$ is the conditional distribution that is estimated by learning the statistical models (e.g., neural networks) to the training data beforehand.
		
		Following the notations in the original paper \citep{huber2014identifying}, let potential outcomes denote $\Yz = Y(0, D(0), M(0))$ and $\Yo = Y(1, D(1), M(0))$, respectively. 
		
		Then given the causal graph in \Cref{fig1}(b), marginal probability $\pr(\Yo = 1)$ can be written as
		\begin{align*}
			&\pr(\Yo = 1) \\
			&= \pr(Y(1, D(1), M(0) ) = 1) \\
			&= \E_{Q}[ \E_{D(1) | Q}[ \E_{M(0) | Q, D(1)} [ \pr ( Y(1, d, m) = 1  | A = 1, Q = q, D(1) = d, M(0) = m  )   ]     ]  ]		.	
			\end{align*}
		Using \Cref{ignorability} in \Cref{sec-a:asmp}, this can be rewritten as
		\begin{align*}
			&\pr(\Yo = 1) = \E_{Q}[ \E_{D | A=1, Q}[ \E_{M | A=0, Q, D} [ \pr( Y(1, d, m) = 1 |  A = 1, Q = q, D = d, M = m    ) ]     ]  ],
			\end{align*}
	With Bayes' theorem, this can be expressed as 
		\begin{align*}
		\pr(\Yo = 1) = \E_{Q}[ \E_{D | Q}[ \E_{M | Q, D} [     \omega'      \pr( Y = 1 | d A = 1, Q = q, D = d, M = m  )  ]     ]  ],
	\end{align*}
	where $\omega'$ is expressed as follows:
	\begin{align*}
		&\omega' = \frac{\pr(A=1 | Q=q, D=d)\pr(A=0 | Q=q, D=d, M=m)}{\pr(A=1 | Q=q) \pr(A=0 | Q=q, D=d)}.
		\end{align*}
	With indicator function $\I(\cdot)$, this can be formulated as
		\begin{align}
		&\pr(\Yo = 1) = \E [   \I(A=1)  w'   \pr( Y = 1 | A = 1, q, d, m  )  ]  \label{mp_huber_exp1} ,
	\end{align}
	where weight $w'$ is expressed as
	\begin{align*}
		w' = \frac{1}{\pr(A=1 | Q=q, D=d, M=m)} \omega'.
		\end{align*}
	
	In a similar manner, marginal probability $\pr(\Yz = 1)$ can be represented as
		\begin{align}
		&\pr(\Yz = 1) = \E [   \I(A=0)  w      \pr( Y = 1 | A = 0, q,  d,  m  )  ]  \label{mp_huber_exp0} ,
	\end{align}
	where weight $w$ is formulated as
	\begin{align*}
		w = \frac{1}{\pr(A=0 | Q=q)}.
	\end{align*}

	Given empirical distribution, by plugging conditional distribution $c_{\theta}$ into $\pr(Y=1 | A=1, Q=q, D=d, M=m)$, we can estimate  \eqref{mp_huber_exp0} and \eqref{mp_huber_exp1} as \eqref{mp_huber} and derive the estimators \eqref{mp_huber}.

	\section{Computation time and convergence guarantee} \label{sec-a:convergence}

	To minimize the objective function \eqref{eq-Opt3}, we use the stochastic gradient descent method \citep{sutskever2013importance}.
	
	With this method, we computed the penalty term and its gradient over the samples in each mini-batch. The computation time, which is required to evaluate the objective function  \eqref{eq-Opt3} and its gradient is as much as the time needed to evaluate the training loss in \eqref{eq-Opt3} and its gradient, respectively.
	
	Whether we can guarantee that the gradient descent method converges depends on the choice of classifier $h_{\theta}$. For instance, if we choose a neural network classifier, we cannot guarantee that the stochastic gradient descent method \citep{sutskever2013importance} converges because the objective function \eqref{eq-Opt3} becomes nonconvex, and its gradient does not become {\it Lipschitz continuous}; that is, the maximum rate of change in the gradient is not bounded. However, if the neural network only contains activation functions whose gradients are Lipschitz continuous (e.g., the sigmoid function), we can optimize the objective function with convergence guarantees using e.g., the gradient sampling method \citep{burke2005robust}, because in this case, the gradient of the objective function becomes {\it locally Lipschitz continuous} \citep[Chapter 2]{ferrera2013introduction}.

	\section{PIU values that satisfy fairness conditions} \label{sec-a:comparison}
	
	As described in \Cref{subsec:feasible}, our method effectively makes potential outcomes take the same value (i.e., $\Yz = \Yo$) while FIO does not. To illustrate this, in this section, we compare the possible PIU values that satisfy the fairness condition of each method.
	
	\subsection{Comparing possible PIU values} \label{subsec-a:possible}
	
	We first introduce several notations. Let the (true) marginal probabilities of potential outcomes be $\mz = \pr(\Yz=1)$ and $\mo = \pr(\Yo=1)$, and the (true) joint probabilities of $(\Yz, \Yo) = (0, 0)$, $(0, 1)$, $(1, 0)$, and $(1, 1)$ be $\pzz$, $\pzo$, $\poz$, and $\poo$. 
		
	With these notations, PIU can be formulated as $\pzo + \poz$, and its lower and upper bound can be expressed using marginal probabilities $\alpha$ and $\beta$ as
	\begin{align}
	|\mz - \mo| \leq \pzo + \poz \le \min\{\mo, 1 - \mz \} + \min\{\mz, 1 - \mo \},
	\label{piu-lbub}
	\end{align}
	which we prove in \Cref{subsec-a:prooflbub}. 
	
	With our method, as presented in \Cref{fig2-constrained}, the marginal probabilities are forced to be $(\mz, \mo)$ $\approx$ $(0, 0)$ or $(1, 1)$. If $\mz$ and $\mo$ satisfy this condition, since both lower and upper bounds in \eqref{piu-lbub} become close to zero, PIU is constrained to almost zero (i.e., $\pzo + \poz \approx 0$).
	
	By contrast, as described in \Cref{subsec:feasible}, FIO always accepts the marginal probabilities $(\mz, \mo)$ $=$ $(0.5, 0.5)$. At this point, the lower and upper bounds in \eqref{piu-lbub} become $0$ and $1$: $0 \leq \pzo + \poz \leq 1$. This implies that it is completely unknown whether the PIU value is high since the joint probabilities are unknown in practice. Therefore, FIO cannot ensure that the potential outcomes take the same value for all individuals, which is insufficient to guarantee individual-level fairness.
	
	To support the above discussion on possible PIU values, in what follows, we prove the lower and upper bound on PIU (i.e., \eqref{piu-lbub}). 
	
	\subsection{Proof of (A19)} \label{subsec-a:prooflbub}
	
	Since we already proved the upper bound in \eqref{piu-ub}, below we derive the lower bound in \eqref{piu-lbub}. Since $\mz$ and $\mo$ are marginal probabilities, we have
	\begin{align*}
	\begin{aligned}
	\poz + \poo = \mz, \quad \pzo + \poo = \mo,
	\end{aligned}
	\end{align*}
	which are equivalent to
	\begin{align*}
	\begin{aligned}
	\poz = \mz - \poo, \quad \pzo = \mo - \poo,
	\end{aligned}
	\end{align*}
	respectively. By summing up both, we have
	\begin{align*}
	\pzo + \poz = \mz + \mo - 2 \poo.
	\end{align*}
	Since joint probability $\poo$ is less than marginal probabilities $\mz$ and $\mo$, we have $\poo \leq \min\{\mz, \mo\}$. Therefore,
	\begin{align}
	& \pzo + \poz \ge \mz + \mo - 2 \min\{\mz, \mo\} = |\mz - \mo|. \label{piu-lb}
	\end{align}
	
	Combined with the upper bound on $\pzo + \poz$ in \eqref{piu-ub}, we prove \eqref{piu-lbub}.

	\section{Addressing latent confounders} \label{sec-a:extension}
	
	This section details how our method can be extended to ensure individual-level fairness when there are unobserved variables called latent confounders. 
	
	As described in \Cref{subsec:extension}, although it is extremely challenging to estimate the marginal potential outcome probabilities in this case, we can sometimes guarantee individual-level fairness using lower and upper bounds on them. If these lower and upper bounds are given, we can achieve this by reformulating the penalty function as follows:
	\begin{align}
		G_{\theta}(\textbf{\textit{x}}_1, \dots \textbf{\textit{x}}_n) =  \eUo (1 - \eLz) + (1 - \eLo) \eUz ,
		\tag{\ref{eq-penalty_ex}}
	\end{align}	
	where $\eLz$ and $\eUz$ are the estimated lower and upper bounds on marginal probability $\pr(\Yz = 1)$, respectively, and $\eLo$ and $\eUo$ are the estimated lower and upper bounds on marginal probability $\pr(\Yo = 1)$, respectively.
	
	In general, we cannot estimate lower and upper bounds $\eLz$, $\eUz$, $\eUz$, and $\eUo$. However, when a certain causal graph is given, we can estimate them from data. In what follows, as an example, we detail the existing estimators from a previous work \citep{miles2017partial}.
	
	\subsection{Example of existing estimators}
	
		\begin{figure}[t]
		\includegraphics[height=3.5cm]{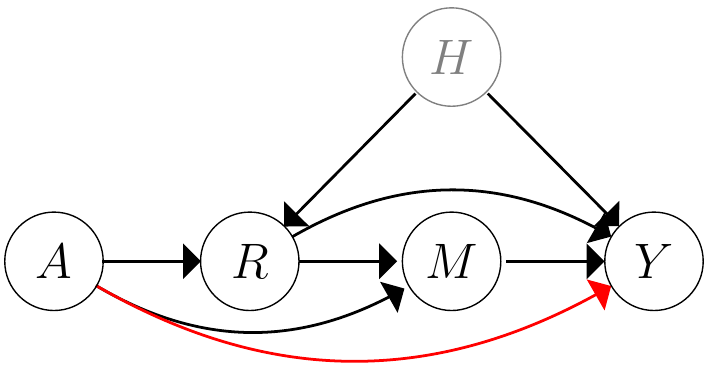}
		\centering 
		\caption{Example of causal graph containing latent confounder $H$ (gray node), which affects both $R$ and $Y$. Red pathway represents unfair pathway $\pi$.} 
		\label{fig-graph}
	\end{figure}
	
	According to a previous result \citep{miles2017partial}, we can estimate the lower and upper bounds on marginal potential outcome probabilities when the causal graph in \Cref{fig-graph} is given. Here $A \in \{0,1\}$ and $Y \in\{0, 1\}$ are binary random variables, $M$ is a discrete one, $R$ can be either a continuous or a discrete one, and $H$ is a latent confounder that affects two variables $R$ and $Y$.\footnote{Regarding $A$, although \citet{miles2017partial} dealt with a general case, where $A$ can be continuous or discrete, we consider a binary case, i.e., $A \in \{0, 1\}$} Here unfair pathway $\pi$ is given as direct pathway $A$ $\rightarrow$ $Y$ (i.e., $\pi = \{A \rightarrow Y\}$). 
	
	With this causal graph, a previous work \citep{miles2017partial} expressed lower and upper bounds on $\pr(\Yo = 1)$ as
	\begin{align*}
	&\hat{l}^{A \Leftarrow 1 \parallel \pi } = \sum_{m} \max\{0,\ \pr(M=m|A=0) - 1 + \sum_{r} \pr(Y=1|A=1, m, r) \pr(R=r|A=1)\},\\
	&\hat{u}^{A \Leftarrow 1 \parallel \pi} = \sum_{m} \min \{\pr(M=m|A=0),\ \sum_{r} \pr(Y=1|A=1, m, r) \pr(R=r|A=1)\}.
	\end{align*}
	With conditional distribution $c_{\theta}(1, M, R) = \pr(Y=1|A=1, M, R)$ provided by classifier $h_{\theta}$, these bounds can be estimated as the following functions of classifier parameter $\theta$: 
	\begin{align}
	&\eLo = \sum_{m} \max\{0,\ \hat{\pr}(M=m|A=0) - 1  + \sum_{r} c_{\theta}(1, m, r) \hat{\pr}(R=r|A=1)\} \label{eq-lower_} \\
	&\eUo = \sum_{m} \min \{\hat{\pr}(M=m|A=0), \ \sum_{r} c_{\theta}(1, m, r) \hat{\pr}(R=r|A=1)\}, \label{eq-upper_}
	\end{align}
	respectively. Here conditional distributions $\hat{\pr}(M=m|A=0)$ and $\hat{\pr}(R=r|A=1)$ can be estimated by learning statistical models (e.g., logistic regression or neural networks) from the training data beforehand.

	As with \eqref{eq-lower_} and \eqref{eq-upper_}, we can formulate the estimated lower and upper bounds on marginal probability $\pr(\Yz = 1)$ as
	\begin{align}
	&\eLz = \sum_{m} \max\{0,\ \hat{\pr}(M=m|A=0) - 1  + \sum_{r} c_{\theta}(0, m, r) \hat{\pr}(R=r|A=0)\}
	\label{lower-e} \\
	&\eUz = \sum_{m} \min \{\hat{\pr}(M=m|A=0), \ \sum_{r} c_{\theta}(0, m, r) \hat{\pr}(R=r|A=0)\},
 \label{upper-e}
	\end{align}
	respectively.
	
	Note that if we use the above lower and upper bounds, solving the optimization problem with convergence guarantees becomes complicated because these lower and upper bounds in the penalty term are not differentiable. Our future work will leverage other bounds on marginal probabilities to formulate an objective function that is differentiable and easy to optimize.

	\section{Experimental settings} \label{sec-a:exp}
	
	This section details the experimental settings presented in \Cref{sec:experiment}.
	
	\subsection{Settings of each method} \label{subsec-a:set}
	
	For classifier $h_{\theta}$ for {\bf Proposed}, {\bf FIO}, {\bf Unconstrained}, and {\bf Remove}, we used a feed-forward neural network that consists of two linear layers with $100$ and $50$ hidden neurons, respectively. We used sigmoid activation functions and formulated the output layer by a log softmax function. For the loss function, we used cross-entropy loss. To train this classifier, we set the minibatch size to $1,000$ for the synthetic data and the Adult dataset. With the German credit dataset, since the number of training samples is less than $1,000$, we set it to $100$. We stopped the training after $1,000$ epochs.

	Unlike these methods, {\bf PSCF} use multiple neural networks to learn a predictive model for each variable in $\{\textbf{\textit{X}}, Y\}$ (see, \Cref{subsubsec-a:synthpscf} for the details of these predictive models). We use the same network architecture as that of the original paper \citep{chiappa2018path}. 
	
	To compare the best performances, we selected the hyperparameter values of {\bf Proposed}, {\bf FIO}, and {\bf PSCF}. For each method, we used a grid search with $0.05$ grid size to select the value of the penalty parameter from [$0.0, 2.0$].
	
	\subsection{Settings in synthetic data experiments} \label{subsec-a:synth}

	\subsubsection{Data} \label{subsubsec-a:synthdata}
	
	We prepared synthetic data that represent a scenario of hiring decisions for physically demanding jobs, whose causal graph is shown in \Cref{fig1}(b). We sampled gender $A \in \{0, 1\}$, qualification $Q$, number of children $D$, physical strength $M$, and hiring decision outcome $Y \in \{0, 1\}$ from the following SEM:
	\begin{align}
	\begin{aligned}
	&A = U_A, \quad  U_A \sim \mathrm{Bernoulli}(0.6), \\
	&Q = \lfloor U_Q \rfloor, \quad U_Q \sim \mathcal{N}(2, 5^2), \\
	&D = A + \lfloor 0.5 Q U_D  \rfloor, \quad U_D \sim \mathrm{Tr}\mathcal{N}(2, 1^2, 0.1, 3.0),\\
	&M = 3 A +  0.4 Q U_M, \quad U_M \sim \mathrm{Tr}\mathcal{N}(3, 2^2, 0.1, 3.0), \\
	&Y = h(A, Q, D, M),
	\end{aligned}\label{eq-synth}
	\end{align}
	where $\mathrm{Bernoulli}$, $\mathcal{N}$, and $\mathrm{Tr}\mathcal{N}$ represent the Bernoulli, Gaussian, and truncated Gaussian distributions, respectively, and $\lfloor\cdot\rfloor$ is a floor function that returns an integer by removing the decimal places. To output hiring decision outcome $Y$, we used function $h$, which is a logistic regression model that provides the following conditional distribution:
	\begin{align*}
	&\pr(Y=1|A, Q, M) = \mathrm{Bernoulli}(\varsigma( -10 + 5A + Q + D + M) ),
	\end{align*}
	where $\varsigma(x)$ $=$ $1 / (1 + \mathrm{exp}(-x))$ is a standard sigmoid function. 
	
	Note that in \eqref{eq-synth}, the structural equations over $D$ and $M$ are not expressed by additive noise models \citep{hoyer2009nonlinear} because they contain multiplicative noises $U_D$ and $U_M$.
	
	In experiments, we used $5,000$ samples to train the classifier and $1,000$ samples to test the performance.	
	
	\subsubsection{Computing unfair effects}  \label{subsubsec-a:synthunfair}
	
	With such synthetic data, we computed the four statistics of unfair effects for {\bf Proposed}, {\bf FIO}, and {\bf Unconstrained} as follows.
	
	To compute (\myra) the mean unfair effect and (\myrc) the upper bound on PIU, we estimated marginal potential outcome probabilities $\ePz$ and $\ePo$ with estimators \eqref{mp_huber}.
	
	To obtain (\myrb) the standard deviation in the conditional mean unfair effects and (\myrd) the PIU value, we sampled potential outcomes $\Yz$ and $\Yo$ based on the SEM \eqref{eq-synth}. Specifically, we sampled $\Yz$ from the following (interventional) SEM:
	\begin{align*}
	\begin{aligned}
	&A = 0, \\
 &Q = \lfloor U_Q \rfloor, \quad U_Q \sim \mathcal{N}(2, 5^2), \\
 &D(0) = \lfloor 0.5 Q U_D  \rfloor, \quad U_D \sim \mathrm{Tr}\mathcal{N}(2, 1^2, 0.1, 3.0),\\
 &M(0) =  0.4 Q U_M, \quad U_M \sim \mathrm{Tr}\mathcal{N}(3, 2^2, 0.1, 3.0), \\
	&\Yz = h_{\theta}(0, Q, D(0), M(0)),
	\end{aligned}
	\end{align*}
	where $h_{\theta}$ is the classifier. We sampled $\Yo$ from
	\begin{align*}
	\begin{aligned}
	&A = 1, \\
	&Q = \lfloor U_Q \rfloor, \quad U_Q \sim \mathcal{N}(2, 5^2), \\
	&D(1) = 1 + \lfloor 0.5 Q U_D  \rfloor, \quad U_D \sim \mathrm{Tr}\mathcal{N}(2, 1^2, 0.1, 3.0),\\
	&M(0) =  0.4 Q U_M, \quad U_M \sim \mathrm{Tr}\mathcal{N}(3, 2^2, 0.1, 3.0), \\
	&\Yo = h_{\theta}(1, Q, D, M).
	\end{aligned}
	\end{align*}
	
	Then using $n$ pairs of these samples $\{(y_{A\Leftarrow 0, i}, y_{A \Leftarrow 1 \parallel \pi, i})\}_{i=1}^n$, we evaluated the PIU value by
	\begin{align*}
	\hat{\pr}(\Yz \neq \Yo) = \frac{1}{n} \sum_{i=1}^n \I( y_{A\Leftarrow 0, i} \neq y_{A \Leftarrow 1 \parallel \pi, i}   ),
	\end{align*}
	where $\I(\cdot)$ is an indicator function that takes $1$ if $y_{A\Leftarrow 0, i} \neq y_{A \Leftarrow 1 \parallel \pi, i} $ ($i$ $\in$ $\{1, \dots, n\}$) and $0$ otherwise. 
	
	We computed the standard deviation in conditional mean unfair effects as follows. We separated $n$ individuals into $K$ subgroups of individuals who have the same values of features $\textbf{\textit{X}}$, took the mean unfair effects over individuals in each subgroup, and computed their standard deviation. Let the individuals in the $k$-th subgroup ($k = 1, \dots, K$) have identical feature attributes $\textbf{\textit{X}}$ $=$ $\textbf{\textit{x}}^k$, where superscript $k$ represents the $k$-th subgroup.  Then using  $\{(y_{A\Leftarrow 0, i}, y_{A \Leftarrow 1 \parallel \pi, i})\}_{i=1}^n$, we estimated the standard deviation of the conditional mean unfair effects over the $K$ subgroups as
	\begin{align}
	\hat{\sigma} = \sqrt{ \frac{  \sum_{k=1}^K \hat{\mu}^k - \hat{\mu} }{K}   }. \label{condmean}
	\end{align}
	Here $\hat{\mu}^k$ is the estimated conditional mean unfair effect in the $k$-th subgroup of individuals with identical attributes $\textbf{\textit{X}}$ $=$ $\textbf{\textit{x}}^k$, i.e.,
	\begin{align*}
	\hat{\mu}^k = \frac{1}{n^k} \sum_{i \in \{1, \dots, n\} | \textbf{\textit{x}}_i = \textbf{\textit{x}}^k}  \I(y_{A\Leftarrow 0, i}\neq y_{A \Leftarrow 1 \parallel \pi, i}),
	\end{align*}
	where $n^k$ is the number of individuals in the $k$-th subgroup and $\hat{\mu}$ is the mean of $\hat{\mu}^k$ over $k=1, \dots, K$, i.e.,
	\begin{align*}
	\hat{\mu} = \frac{1}{K} \sum_{k}^K \hat{\mu}^k.
	\end{align*}
	
	\subsubsection{Unfair effects of  PSCF}  \label{subsubsec-a:synthpscf}
	
	With {\bf PSCF}, we did not evaluate the two statistics of unfair effects, (\myra) the mean unfair effect and (\myrc) the upper bound on PIU, since they are not well-defined for this method. 
	
	This is because these statistics measure the unfairness of the learned predictive model of $Y$ (i.e., classifier $h_{\theta}$ in our method); however, {\bf PSCF} aims to ensure fairness using {\it unfair} predictive models. 
	
	To do so,  {\bf PSCF} approximates the SEM by learning predictive models of each variable in $\pr(\textbf{\textit{X}}, Y)$, which is unfair due to the discriminatory bias in the observed data, and removes the unfairness by sampling the {\it fair} feature values based on the approximated SEM.
	
	For instance, in the case of synthetic data experiments, {\bf PSCF} approximates the SEM \eqref{eq-synth} as follows. Using latent variable $H_D$, {\bf PSCF} learns the predictive models of $A$, $Q$, $D$, $M$, and $Y$ and the distribution of $H_D$, which are expressed as follows:
	\begin{align}
		\begin{aligned}
			&A = \pr_{\theta}(A), \\
			&Q = \pr_{\theta}(Q), \\
			&D = \pr_{\theta}(D | A, Q, H_D), \quad H_D = \pr_{\theta}(H_D),\\
			&M = \pr_{\theta}(M | A, Q),\\
			&Y = \pr_{\theta}(Y | A, Q, D, M),
		\end{aligned}\label{eq-pscf}
	\end{align}
	where $\pr_{\theta}$ denotes a (conditional) distribution, which is parameterized as a neural network. Here latent variable $H_D$ approximates the additive noise in the structural equation over $D$.
		
	 By using the approximated SEM \eqref{eq-pscf}, {\bf PSCF} aims to make fair predictions as follows. For each individual $i$ $\in$ $\{1, \dots, n\}$ with attributes $\{a_i, q_i, d_i, m_i\}$, {\bf PSCF} samples their {\it fair} attribute of $D$ by
	 	\begin{align}
	 		\hat{d}(0)_i \sim \pr_{\theta}(D | A=0, q_i, h_{D, i}), \quad h_{D, i} \sim \pr_{\theta}(H_D).
	\label{eq-pscfpm}
	 \end{align}
	Using this attribute, {\bf PSCF} makes a prediction using the following Monte Carlo estimate:
	 	\begin{align}
		&\hat{y}^{PSCF}_i = \frac{1}{J} \sum_{j=1}^J \hat{y}^{PSCF}_{i, j} \quad \text{where}\quad \hat{y}^{PSCF}_{i, j} \sim  \pr_{\theta}(Y | A=0, q_i, d(0)_i, m_i).
		\label{eq-pscfpred}
	\end{align}
	Here $J$ is the number of Monte Carlo samples, which is set to $J=5$ in our experiments. 
	
	According to \citet{chiappa2018path}, if there is slight mismatch between the approximated SEM \eqref{eq-pscf} and the true SEM \eqref{eq-synth}, {\bf PSCF} can eliminate the conditional mean unfair effect and achieve individual-level fairness. Intuitively, this is because in \eqref{eq-pscfpred} and \eqref{eq-pscfpm}, $A$'s values are fixed in the approximated structural equations over $Y$ and $D$, which involve unfair pathways $\pi = \{A \rightarrow Y, A \rightarrow D \rightarrow Y\}$.
	
	In this way, {\bf PSCF} aims to achieve fairness by sampling the fair feature values using unfair predictive models. Therefore, we cannot measure the unfairness of this method using the two statistics (\myra) and (\myrc), which measure the unfairness of the predictive model.
	
	In contrast, it is appropriate to measure the unfairness based on two other statistics, i.e., (\myrb) the standard deviation in the conditional mean unfair effects and (\myrd) the PIU value. Since both are formulated using the true SEM, they can be used to quantify the unfairness due to SEM's approximation error.
		
	To compute the unfairness of {\bf PSCF} using these two statistics, we made a prediction in the same way as \eqref{eq-pscfpred}, except that we used the true SEM \eqref{eq-synth}. Let such predicted values be $\{y^{PSCF}_i\}_{i=1}^n$. Then using $n$ pairs of predicted values $\{(y^{PSCF}_{i}, \hat{y}^{PSCF}_{i})\}_{i=1}^n$, we estimated (\myrd) the PIU value as
	\begin{align*}
		\hat{\pr}(\Yz \neq \Yo) = \frac{1}{n} \sum_{i=1}^n \I( y^{PSCF}_{i} \neq \hat{y}^{PSCF}_{i}   ),
	\end{align*}
	and (\myrb) the standard deviation of conditional mean unfair effects in the same way as \eqref{condmean} except that $\hat{\mu}^k$ is estimated as
	\begin{align*}
		\hat{\mu}^k = \frac{1}{n^k} \sum_{i \in \{1, \dots, n\} | \textbf{\textit{x}}_i = \textbf{\textit{x}}^k}  \I(y^{PSCF}_{i}\neq \hat{y}^{PSCF}_{i}).
		\end{align*}

	\subsection{Settings in real-world data experiments} \label{subsec-a:real}
	
	\subsubsection{Data and causal graphs} \label{subsubsec-a:realdata}
	
	In real-world data experiments, we used two datasets, the German credit and Adult datasets \citep{uci}. 
	
	The German credit dataset consists of the records of $1,000$ loan applicants and includes the attributes of each individual, such as gender, amount of savings, and age.  
	
	Using this dataset, we predicted whether each loan applicant is a good or bad credit risk. We used $900$ samples for training and $100$ samples to test the performance. 
	
	To evaluate the unfairness of the predictions, following \cite{chiappa2018path}, we used the causal graph in \Cref{fig2}(a). Here, $A$ denotes gender, $Y$ expresses credit risk, $\textbf{\textit{S}}$ represents financial information (i.e., amount, checking account balance, and house ownership), $\textbf{\textit{R}}$ stands for credit amount and repayment duration, and $\textbf{\textit{C}}$ corresponds to such attributes of each individual as age and loan purpose. We regarded gender $A$ as a sensitive feature and pathways $A \rightarrow Y$ and $A \rightarrow \textbf{\textit{S}} \rightarrow Y$ as unfair.
		
	On the other hand, the Adult dataset is comprised of US census data that contain the features of individuals including gender, occupation, and income. 
	
	With this dataset, we predicted whether each individual has an annual income exceeding $50,000$ US dollars. We employed $34,001$ samples to train our classifier and $10,870$ samples to test the performance.
	
	 Following \cite{chiappa2018path}, we used the causal graph in Fig. \ref{fig2}(b). We regarded gender $A$ as a sensitive feature. Other features are marital status $M$, education $L$, occupation information ${\bf R}$ (e.g., weekly working hours), age and nationality ${\bf C}$, and predicted income $Y$. Unfair pathways $\pi$ are direct pathway $A$ $\rightarrow$ $Y$ and pathways from $A$ to $Y$ through $M$ (i.e., $A$ $\rightarrow$ $M$ $\rightarrow$ $\cdots$ $\rightarrow$ $Y$).
	
	\subsubsection{Computing unfair effects}  \label{subsubsec-a:realunfair}
	
	To evaluate the two statistics of unfair effects (i.e., (\myra) and (\myrc)), we computed the marginal probabilities of potential outcomes $\ePz$ and $\ePo$ based on the existing estimators \citep{huber2014identifying}.

	With the German credit dataset, using the attributes of $n$ individuals $\{a_i, \textbf{\textit{c}}_i, \textbf{\textit{s}}_i, \textbf{\textit{r}}_i\}_{i=1}^n$, we estimated the marginal probabilities as
	\begin{align}
\ePz = \frac{1}{n} \sum_{i=1}^n \I(a_i = 0) \hat{w}_i c_{\theta}(a_i, \textbf{\textit{c}}_i, \textbf{\textit{s}}_i, \textbf{\textit{r}}_i), \quad \ePo = \frac{1}{n} \sum_{i=1}^n \I(a_i = 1) \hat{w}_i' c_{\theta}(a_i, \textbf{\textit{c}}_i, \textbf{\textit{s}}_i, \textbf{\textit{r}}_i),
	\label{mp_german}
\end{align}
respectively, where weights $\hat{w}_i$ and $\hat{w}_i'$ are expressed as
\begin{align*}
	\hat{w}_i = \frac{1}{\hat{\pr}(A=0 | \textbf{\textit{c}}_i)}, \quad \hat{w}_i' = \frac{\hat{\pr}(A=1 | \textbf{\textit{c}}_i, \textbf{\textit{s}}_i) \hat{\pr}(A=0 | \textbf{\textit{c}}_i, \textbf{\textit{s}}_i, \textbf{\textit{c}}_i)}{\hat{\pr}(A=1 |  \textbf{\textit{c}}_i) \hat{\pr}(A=0 |  \textbf{\textit{c}}_i, \textbf{\textit{s}}_i) \hat{\pr}(A=1 |  \textbf{\textit{c}}_i, \textbf{\textit{s}}_i, \textbf{\textit{c}}_i)},
\end{align*}
	To compute these weights, we inferred the conditional probabilities of $A$ by fitting the logistic regression model to the training data beforehand.
	
	For the Adult dataset, given the attributes of $n$ individuals $\{a_i, m_i, l_i, \textbf{r}_i, \textbf{\textit{c}}_i\}_{i=1}^n$, we estimated the marginal probabilities as
	\begin{align*}
	\ePz = \frac{1}{n} \sum_{i=1}^n \I(a_i = 0) \hat{w}_i   c_{\theta}(0, m_i, l_i, \textbf{\textit{r}}_i, \textbf{\textit{c}}_i),  \quad \ePo = \frac{1}{n} \sum_{i=1}^n \I(a_i = 1) \hat{w}'_i c_{\theta}(1, m_i, l_i, \textbf{\textit{r}}_i, \textbf{\textit{c}}_i),
	\end{align*}
	where weights $\hat{w}_i$ and $\hat{w}_i'$ are provided by 
	\begin{align*}
	\hat{w}_i= \frac{1}{\hat{\pr}(A=0|\textbf{\textit{c}}_i)}, \quad 
	\hat{w}'_i = \frac{\hat{\pr}(A=1|m_i, \textbf{\textit{c}}_i) \hat{\pr}(A=0|m_i,l_i, \textbf{\textit{r}}_i, \textbf{\textit{c}}_i) }
	{\hat{\pr}(A=1|\textbf{\textit{c}}_i)  \hat{\pr}(A=0|m_i, \textbf{\textit{c}}_i) \hat{\pr}(A=1|m_i,l_i, \textbf{\textit{r}}_i, \textbf{\textit{c}}_i) }.
	\end{align*}
	To obtain these weight values, we estimated each conditional probability of $A$ by fitting the logistic regression model to the data beforehand.

	\subsection{Computing infrastructure} \label{sec-a:ci}
	
	In our experiments, we used PyTorch 1.6.0 as an implementation of the optimization algorithm \citep{sutskever2013importance} and a 64-bit CentOS machine with 2.6GHz Xeon E5-2697A-v4 (x2) CPUs and 512-GB RAM.

	\section{Additional experimental results} \label{sec-a:addexp}
	
	To further demonstrate the effectiveness of our proposed method, in this section, we provide several additional experimental results. 
	
	This section is organized as follows. In \Cref{subsec-a:logisticexp}, we show that our method works well even with a simpler classifier than the neural network used in \Cref{sec:experiment}. In \Cref{subsec-a:errbexp}, we investigate the statistical significance of the test accuracy using the test set bound \citep{langford2005tutorial}. In \Cref{subsec-a:admexp}, we confirm that when the data satisfy the functional assumptions of the PSCF method \citep{chiappa2018path}, both our method and PSCF achieve individual-level fairness. In \Cref{subsec-a:boundexp}, we show the tightness of the upper bound on PIU by confirming that the performance does not differ so much even when having an oracle access to the true PIU value. Finally, in \Cref{subsec-a:extendedexp}, we evaluate the performance of our extended framework described in \Cref{sec-a:extension}.
			
	\subsection{Results with logistic regression model} \label{subsec-a:logisticexp}
	
	In \Cref{sec:experiment}, we evaluated the performance of our method with a two-layered feed-forward neural network as a classifier. To confirm that our method also works well with a simpler classifier, we show its experimental results with logistic regression model. 
	
	We compared the performance of our method ({\bf Proposed}) with three baselines: {\bf FIO} \citep{nabi2018fair}, {\bf Unconstrained}, and {\bf Remove} \citep{kusner2017counterfactual}. Here, we did not use {\bf PSCF} \citep{chiappa2018path} as a baseline because it is a neural network-based model.
	
	\Cref{table_logistic} and \Cref{fig-ue_logistic} present the test accuracy and the four statistics of unfair effects. As with the results with the neural network shown in \Cref{sec:experiment}, the unfair effects of {\bf Proposed} were much closer to zero than {\bf FIO} and {\bf Unconstrained}, and its test accuracy exceeded {\bf Remove}. These experimental results indicate that our method can achieve a good performance without using a complex classifier such as a neural network.
	
	\begin{table}[t] 
		\centering
		\caption{Test accuracy (\%) on each dataset when using logistic regression model}
		\label{table_logistic}
			\tabcolsep=0.5mm
			\begin{tabular}{lccc}
				\toprule
				Method & Synth & German & Adult \\		
				\midrule						
				{\bf Proposed}&     78.2 $\pm$ 1.5  & 76.0 &   75.2  \\
				{\bf FIO} &     	83.4 $\pm$ 1.2  & 77.5  &  79.0 \\
				{\bf Unconstrained}  & 87.8 $\pm$ 0.8  & 78.8  &  79.7  \\
				{\bf Remove}  & 76.1 $\pm$ 0.9  & 73.8  &  74.4  \\
				\bottomrule
			\end{tabular}
	\end{table}

		\begin{figure}[t]
		\includegraphics[height=7cm]{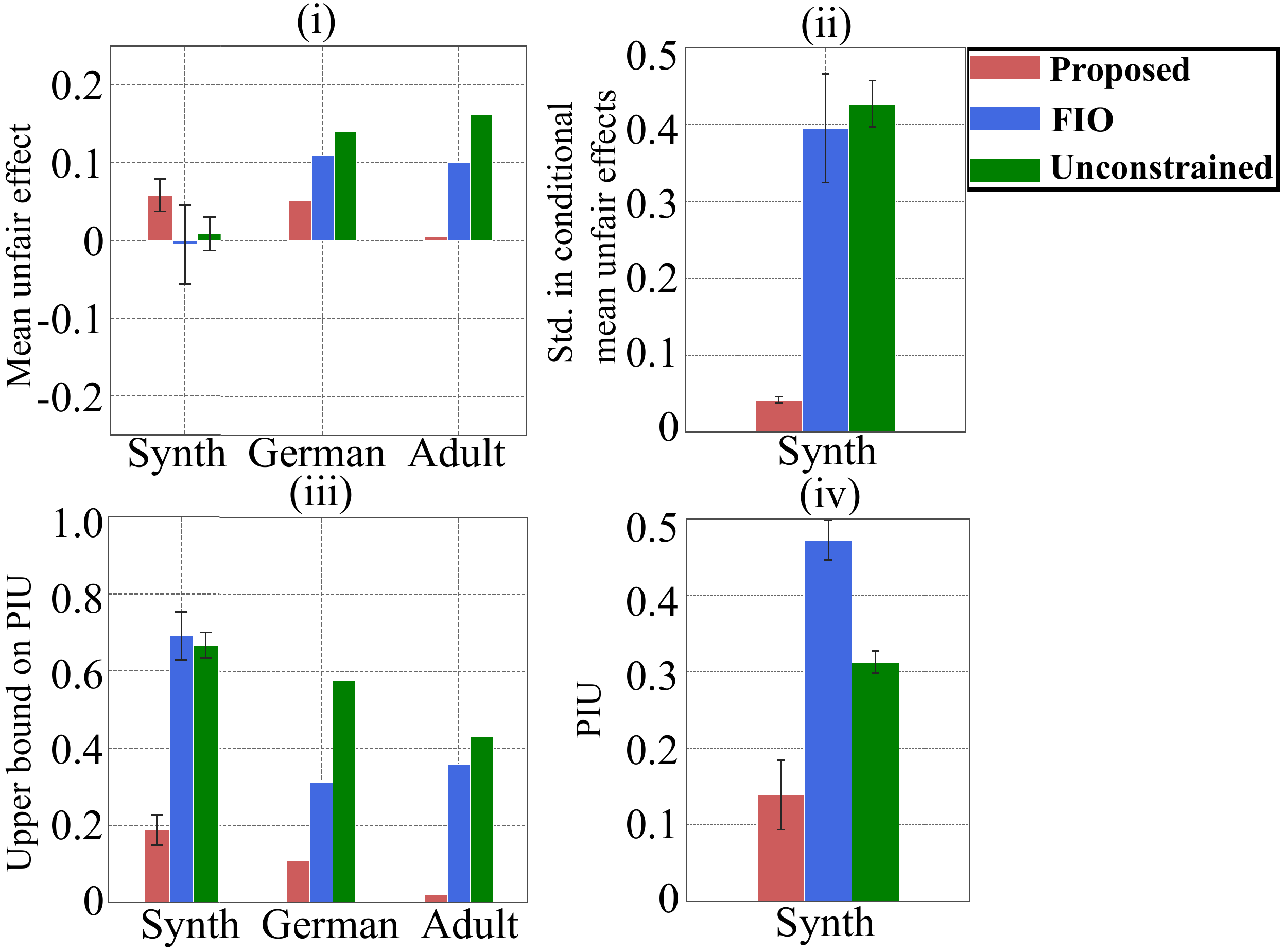}
		\centering 
		\caption{Four statistics of unfair effects on test data when using logistic regression model. The closer they are to zero, the fairer predictions are. Error bars express standard deviations in 10 runs with different datasets. With {\bf Remove}, all statistics are zero (not shown).} 
		\label{fig-ue_logistic}
	\end{figure}

	\subsection{Confidence intervals on error rates on real-world data} \label{subsec-a:errbexp}
	
		\begin{table}[t] 
		\centering
		\caption{Test error and test set bound (\%) on each real-world dataset when using logistic regression model.}
		\label{table_testsetbound}
		\tabcolsep=0.5mm
		\begin{tabular}{lcccc}
			\toprule
			\multirow{2}{*}{Method} & \multicolumn{2}{c}{German}  & \multicolumn{2}{c}{Adult}   \\
			& Test error & Test set bound & Test error & Test set bound \\
			\midrule
			{\bf Proposed}  &  24.0   & [18.0, 32.1]   & 24.8 & [24.1, 25.5]               \\
			{\bf FIO}  &  22.5   & [17.1, 31.0]   & 21.0 & [20.4, 21.7]             \\
			{\bf Unconstrained}  &  21.2   & [15.4, 28.8]   & 20.3 & [19.7, 20.9]               \\
			{\bf Remove}  &  26.2   & [19.8, 34.2]   & 26.6 & [25.9, 27.3]             \\
			\bottomrule
			
		\end{tabular}
	\end{table}

	In \Cref{table1,table_logistic}, we present the test accuracy of each method. These results contain the standard deviations on synthetic datasets, which are computed using randomly generated data; however, they do not include those on real-world datasets. For this reason, in this section, we evaluated the statistical significance of the test accuracy on real-world datasets.
	
	We computed the confidence interval of the test error using the test set bound \citep{langford2005tutorial} (with error rate $\delta = 0.05$), which is a widely-used error measure for binary classification. Here, as with \Cref{subsec-a:logisticexp}, we used logistic regression model as a classifier.
	
		\begin{table}[H] 
		\centering
		\caption{Test accuracy (\%) on synthetic data that satisfy functional assumptions of PSCF method}
		\label{table_adm}
		\tabcolsep=0.5mm
		\begin{tabular}{lc}
			\toprule
			Method & Test accuracy (\%) \\
			\midrule
			{\bf Proposed} & 72.5 $\pm$ 1.1 \\			
			{\bf PSCF}  & 72.5 $\pm$ 0.4  \\
			{\bf Unconstrained} & 80.0 $\pm$  1.3  \\
			\bottomrule
		\end{tabular}
	\end{table}
	
	\begin{figure}[H]
		\includegraphics[height=3cm]{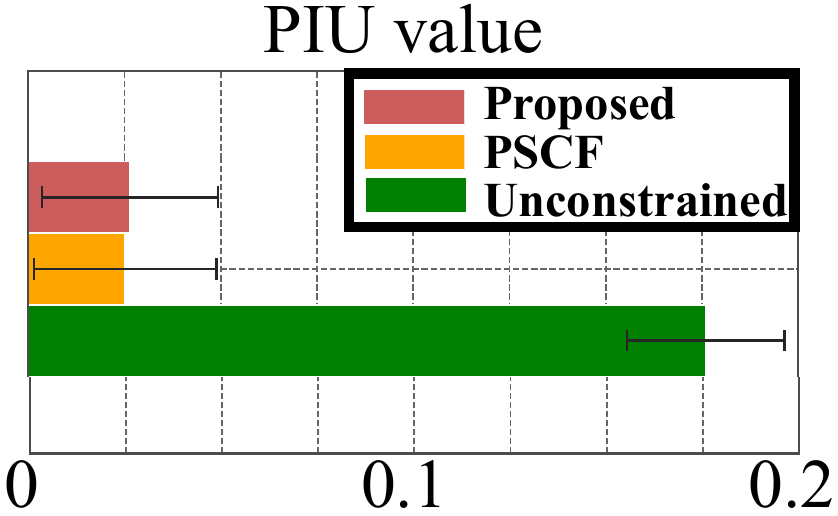}
		\centering 
		\caption{PIU values on test data. The closer they are to zero, the fairer predictions are. Error bars express standard deviations in 10 runs with randomly generated datasets.} 
		\label{fig_adm}
	\end{figure}

	\Cref{table_testsetbound} shows the results. With the German dataset, the test errors did not significantly differ due to the small sample size. However, with the Adult dataset, {\bf Proposed} achieved a significantly lower test error than {\bf Remove} and kept the unfair effect values close to zero (\Cref{fig-ue_logistic}). Thus, we confirmed the statistical significance of the test accuracy on the Adult dataset.

	\subsection{Performance on data that satisfy functional assumptions} \label{subsec-a:admexp}
	
	In \Cref{sec:experiment}, using synthetic data that do not satisfy the functional assumptions of the PSCF method, we show that when those assumptions are violated, PSCF cannot achieve individual-level fairness. By contrast, in this section, we provide experimental results using synthetic data that satisfy the functional assumptions and confirm that with such data, both our method and PSCF can learn an individually fair classifier.	
	
	\subsubsection{Data}
	
	Since the PSCF method assumes that the data are generated by additive noise models \citep{hoyer2009nonlinear}, following this assumption, we prepared the synthetic data. We prepared $5,000$ training samples and $1,000$ test samples using the following additive noise model:
		\begin{align}
		\begin{aligned}
			&A = U_A, \quad  U_A \sim \mathrm{Bernoulli}(0.6), \\
			&Q = \lfloor U_Q \rfloor, \quad U_Q \sim \mathcal{N}(2.5, 5^2), \\
			&D = A + \lfloor 0.1 Q \rfloor + \lfloor U_D \rfloor, \quad U_D \sim \mathcal{N}(1, 0.5^2),\\
			&M = 3 A +  \lfloor0.4 Q \rfloor + \lfloor U_M \rfloor, \quad U_M \sim \mathcal{N}(1, 0.5^2), \\
			&Y = h(A, Q, D, M).
		\end{aligned}\label{eq-additive}
	\end{align}

	\subsubsection{Results}

	With the above data, we compared the performance of {\bf Proposed} with {\bf PSCF} and {\bf Unconstrained}. 
	
	\Cref{table_adm} and \Cref{fig_adm} show the test accuracies and the PIU values. The test accuracies of {\bf Proposed} and {\bf PSCF} were almost the same, which were lower than {\bf Unconstrained}. Their PIU values were much close to zero than {\bf Unconstrained}. These results demonstrate that if the data satisfy the functional assumptions, {\bf Proposed} and {\bf PSCF} can achieve almost the same performance and make individually fair predictions.

	\subsection{Effectiveness of proposed upper bound on PIU}  \label{subsec-a:boundexp}
	
	To demonstrate the tightness of our upper bound on PIU, we compared our {\bf Proposed} with {\bf Oracle}, which uses true PIU values as penalties during the training phase with the same penalty parameter value. As with the experiments presented in \Cref{sec:experiment}, we used a two-layered neural network as a classifier.
	
	\Cref{table_bound} presents the test accuracy and the PIU value on the synthetic dataset. Both the test accuracy and the PIU value did not differ much, even if we have an oracle access to the true PIU values. These results show the effectiveness of our upper bound on PIU.
	
	\begin{table}[t] 
	\centering
	\caption{Test accuracy and PIU value on synthetic data}
	\label{table_bound}
	\tabcolsep=0.5mm
	\begin{tabular}{lcc}
		\toprule
		Method & Test accuracy (\%) & PIU ($\times 10^{-2}$)  \\		
		\midrule						
		{\bf Proposed}&     80.0 $\pm$ 0.9  & 5.04 $\pm$ 3.25   \\
		{\bf Oracle} &     	78.7 $\pm$ 0.9  & 2.63  $\pm$  0.90 \\
		\bottomrule
	\end{tabular}
\end{table}

	\subsection{Evaluating extended framework} \label{subsec-a:extendedexp}

	In this section, we show the empirical performance of our extended framework ({\bf Proposed}$_{ex}$), which addresses cases with latent confounders (see \Cref{sec-a:extension} for the details of our extended framework).

	\subsubsection{Experimental settings} \label{subsubsec-a:extendedexpset}
	
	We evaluated the performance with synthetic data that contain a latent confounder. Following the causal graph in \Cref{fig-graph}, we prepared such data by sampling from the following SEM:
	\begin{align}
	\begin{aligned}
	&A = U_A, \quad  U_A \sim \mathrm{Bernoulli}(0.6), \\
	&R = 3 A + \lfloor 10 H \rfloor + \lfloor U_R \rfloor, \quad U_R \sim \mathcal{N}(1, 0.5^2),\\
	&M = A + R + \lfloor U_M \rfloor, \quad U_M \sim \mathcal{N}(1, 0.5^2), \\
	&Y = h(A, R, M, H),
	\end{aligned}\label{eq-synth-extended}
	\end{align}
	where $H$ denotes a latent confounder, which is sampled by $H \sim \mathcal{N}(1, 0.5^2)$, and function $h$ expresses a logistic regression model that provides the following conditional distribution:
	\begin{align*}
	\pr(Y=1|A, R, M, H) = \mathrm{Bernoulli}(\varsigma( -10 + 5A + R + M + 10 H) ).
	\end{align*}
	
	Given such a synthetic dataset, we used $5,000$ samples for training and $1,000$ samples to test the performance. Other settings are given in the same way as \Cref{subsec-a:set}.
	
	To evaluate the unfair effects, we used the causal graph in \Cref{fig-graph} with unfair pathway $\pi = \{A \rightarrow Y\}$. With SEM \eqref{eq-synth-extended}, we computed (\myrb) the standard deviation in the conditional mean unfair effects and (\myrd) the PIU in a similar manner as in the synthetic data experiments in \Cref{sec:experiment}. For a fair comparison, we did not evaluate the other two statistics (i.e., (\myra) and (\myrc)) because they depend on marginal potential outcome probabilities, whose estimators are formulated in different ways between {\bf Proposed}$_{ex}$ and other methods. 

	Note that in this experiment, as with {\bf Remove}, the unfair effects of {\bf PSCF} become exactly zero. This is because in this case, {\bf PSCF} makes prediction $Y$ by fixing $A$'s value to zero for all individuals, which completely removes the unfair effect along $\pi = \{A \rightarrow Y\}$.

	\subsubsection{Results} 
	
		\begin{table}[t] 
		\centering
		\caption{Test accuracy on synthetic data with latent confounder: Results are shown by (mean $\pm$ standard deviation), computed based on $10$ runs with randomly generated different datasets. }
		\label{table4}
		\scalebox{1.0}{
			\tabcolsep=0.5mm
			\begin{tabular}{lccc}
				\toprule
				Method & Test accuracy (\%) \\		
				\midrule					
				{\bf Proposed}$_{ex}$ &  95.7 $\pm$ 0.5    \\	
				{\bf Proposed}&   96.1 $\pm$ 0.6    \\
				{\bf FIO} &     96.3 $\pm$ 0.6 	 \\
				{\bf PSCF} &     93.8 $\pm$ 0.9	 \\
				{\bf Unconstrained}  & 97.2 $\pm$ 0.6   \\
				{\bf Remove}  & 94.0 $\pm$ 0.6  \\
				\bottomrule
			\end{tabular}
		}
	\end{table}
	
	\begin{figure}[t]
		\includegraphics[height=5.5cm]{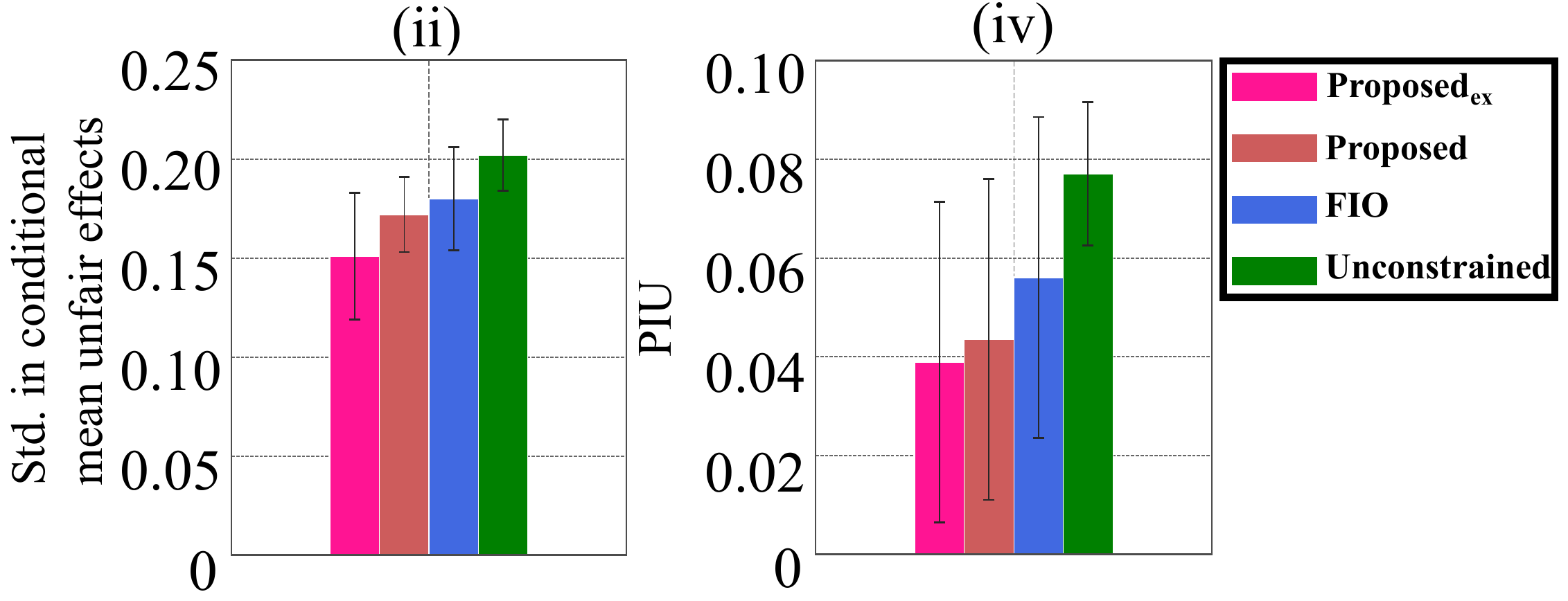}
		\centering 
		\caption{Two statistics of unfair effects on test data. The closer they are to zero, the fairer predictions are. Error bars express standard deviations in 10 runs with randomly generated datasets. As described in \Cref{subsubsec-a:extendedexpset}, with \textbf{PSCF} and \textbf{Remove}, both statistics are zero (not shown).} 
		\label{fig-ue_ex}
	\end{figure}
	
	We present the test accuracy in \Cref{table4} and the unfair effects in \Cref{fig-ue_ex}. The results are shown as the mean and standard deviation in $10$ experiments with randomly generated data. 
		
	With {\bf Proposed}$_{ex}$, both the statistics of the unfair effects were closer to zero than {\bf Proposed} and {\bf FIO} because it uses more reliable estimators of marginal potential outcome probabilities. These results demonstrate that our proposed extension makes fairer predictions than these methods.
	
	The test accuracy of {\bf Proposed}$_{ex}$ exceeded {\bf PSCF} and {\bf Remove}, both of which completely eliminates unfair effects as described in \Cref{subsubsec-a:extendedexpset}, indicating that our {\bf Proposed}$_{ex}$ can strike a better balance between prediction accuracy and fairness than these two methods.
	
	These results imply that with our proposed extension, we can effectively address cases with latent confounders. Although achieving a good balance between accuracy and fairness in such cases remains an open problem, if there are reliable estimators of lower and upper bounds on marginal potential outcome probabilities, our proposed extension will enable us to strike a good balance between individual-level fairness and prediction accuracy.

\end{document}